\documentclass[lettersize,journal]{IEEEtran}
\usepackage{amsmath,amsfonts}
\usepackage{algorithmic}
\usepackage{algorithm}
\usepackage{array}
\usepackage[caption=false,font=normalsize,labelfont=sf,textfont=sf]{subfig}
\usepackage{textcomp}
\usepackage{stfloats}
\usepackage{url}
\usepackage{verbatim}
\usepackage{graphicx}
\usepackage{cite}

\usepackage{multirow}
\usepackage{color}
\usepackage[table,xcdraw]{xcolor}

\usepackage[colorlinks,
            linkcolor=black,
            anchorcolor=black,
            citecolor=black,]{hyperref}
\usepackage{bm}

\definecolor{lightpurple}{RGB}{221,160,221}

\begin{document}

\title{Boundary Discretization and Reliable Classification Network for Temporal Action Detection}

\author{Zhenying Fang, Jun Yu, \textit{Senior Member, IEEE}, and Richang Hong, \textit{Senior Member, IEEE}
\thanks{(Corresponding author: Richang Hong.)}
\thanks{Z. Fang and R. Hong are with the School of Computer and Information Engineering, Hefei University of Technology, 188065, China. e-mail: {zhenyingfang@mail.hfut.edu.cn, hongrc@hfut.edu.cn}.}
\thanks{J. Yu is with the School of Computer Science and Technology, Hangzhou Dianzi University, Hangzhou, 310018, China. e-mail: {yujun@hdu.edu.cn}.}
\thanks{Manuscript received April 19, 2021; revised August 16, 2021.}
}

\markboth{Journal of \LaTeX\ Class Files,~Vol.~14, No.~8, August~2021}%
{Shell \MakeLowercase{\textit{et al.}}: A Sample Article Using IEEEtran.cls for IEEE Journals}


\maketitle

\begin{abstract}
Temporal action detection aims to recognize the action category and determine each action instance's starting and ending time in untrimmed videos. The mixed methods have achieved remarkable performance by seamlessly merging anchor-based and anchor-free approaches. Nonetheless, there are still two crucial issues within the mixed framework: (1) Brute-force merging and handcrafted anchor design hinder the substantial potential and practicality of the mixed methods. (2) Within-category predictions show a significant abundance of false positives. In this paper, we propose a novel Boundary Discretization and Reliable Classification Network (BDRC-Net) that addresses the issues above by introducing boundary discretization and reliable classification modules. Specifically, the boundary discretization module (BDM) elegantly merges anchor-based and anchor-free approaches in the form of boundary discretization, eliminating the need for the traditional handcrafted anchor design. Furthermore, the reliable classification module (RCM) predicts reliable global action categories to reduce false positives. Extensive experiments conducted on different benchmarks demonstrate that our proposed method achieves competitive detection performance. Our source code is available at \url{https://github.com/zhenyingfang/BDRC-Net}.
\end{abstract}



\section{introduction}\label{sec:introduction}
\IEEEPARstart{T}{emporal} Action Detection (TAD) is a fundamental yet challenging task of video understanding, aiming to identify and localize action instances within untrimmed videos. Various applications, such as video summarization \cite{videosummary}, intelligent surveillance \cite{surveillance1}, and human behavior analysis \cite{behavior1, behavior2}, can be feasible with the help of TAD.

The TAD methods can be classified into three types: anchor-based, anchor-free, query-based, and mixed. Anchor-based methods~\cite{ssad, rc3d, pbrnet, tbos, tallformer} rely on the manually defined anchors, and their performance is sensitive to the number and scale of anchors designed. Thanks to the assistance of carefully designed anchors, their prediction stability is better. Anchor-free methods ~\cite{ssn, bsn, afsd, brem, actionformer} densely predict action categories and corresponding boundaries on every snippet of the entire video. Compared with anchor-based methods, they can detect more flexible continuous actions. Query-based methods~\cite{rtdnet, tadtr, react, selfdetr, dualdetr, tetad} introduce a set prediction mechanism to simplify the post-processing of detectors. Recently, mixed methods~\cite{pcad, mgg, a2net} have achieved remarkable results by combining the two methods mentioned above. While maintaining the stability of the anchor-based methods, flexibility has been enhanced by relying on the anchor-free methods.

\begin{table}[t!]
    \caption{Removing false positive category predictions improves the performance on THUMOS'14 of mixed method A2Net and anchor-free method ActionFormer. $^\dagger$: Our reproduce.}
    \centering
    \resizebox{1.0\linewidth}{!}{\begin{tabular}{c|ccccccc}
\hline
\multirow{2}{*}{Method} & \multicolumn{7}{c}{mAP@tIoU(\%)}                   \\
                        & 0.1  & 0.2  & 0.3  & 0.4  & 0.5  & 0.6  & 0.7  \\ \hline
A2Net$^\dagger$                   & 67.9 & 65.6 & 61.8 & 54.3 & 44.5 & 33.0 & 19.1 \\
A2Net\_remove\_fp$^\dagger$       & 73.4 ($\uparrow$ \textcolor{red}{5.5}) & 70.8 ($\uparrow$ \textcolor{red}{5.2}) & 66.7 ($\uparrow$ \textcolor{red}{4.9}) & 58.3 ($\uparrow$ \textcolor{red}{4.0}) & 47.2 ($\uparrow$ \textcolor{red}{2.7}) & 34.5 ($\uparrow$ \textcolor{red}{1.5}) & 19.8 ($\uparrow$ \textcolor{red}{0.7}) \\
ActionFormer                   & 85.4 & 84.5 & 82.1 & 77.8 & 71.0 & 59.4 & 43.9 \\
ActionFormer\_remove\_fp       & 87.8 ($\uparrow$ \textcolor{red}{2.4}) & 86.9 ($\uparrow$ \textcolor{red}{2.4}) & 84.6 ($\uparrow$ \textcolor{red}{2.5}) & 80.0 ($\uparrow$ \textcolor{red}{2.2}) & 73.0 ($\uparrow$ \textcolor{red}{2.0}) & 61.4 ($\uparrow$ \textcolor{red}{2.0}) & 45.6 ($\uparrow$ \textcolor{red}{1.7}) \\ \hline
\end{tabular}}
\label{tab:a2net_remove_fp}
\end{table}

\begin{figure}
\begin{center}
    \includegraphics[width=0.95\linewidth]{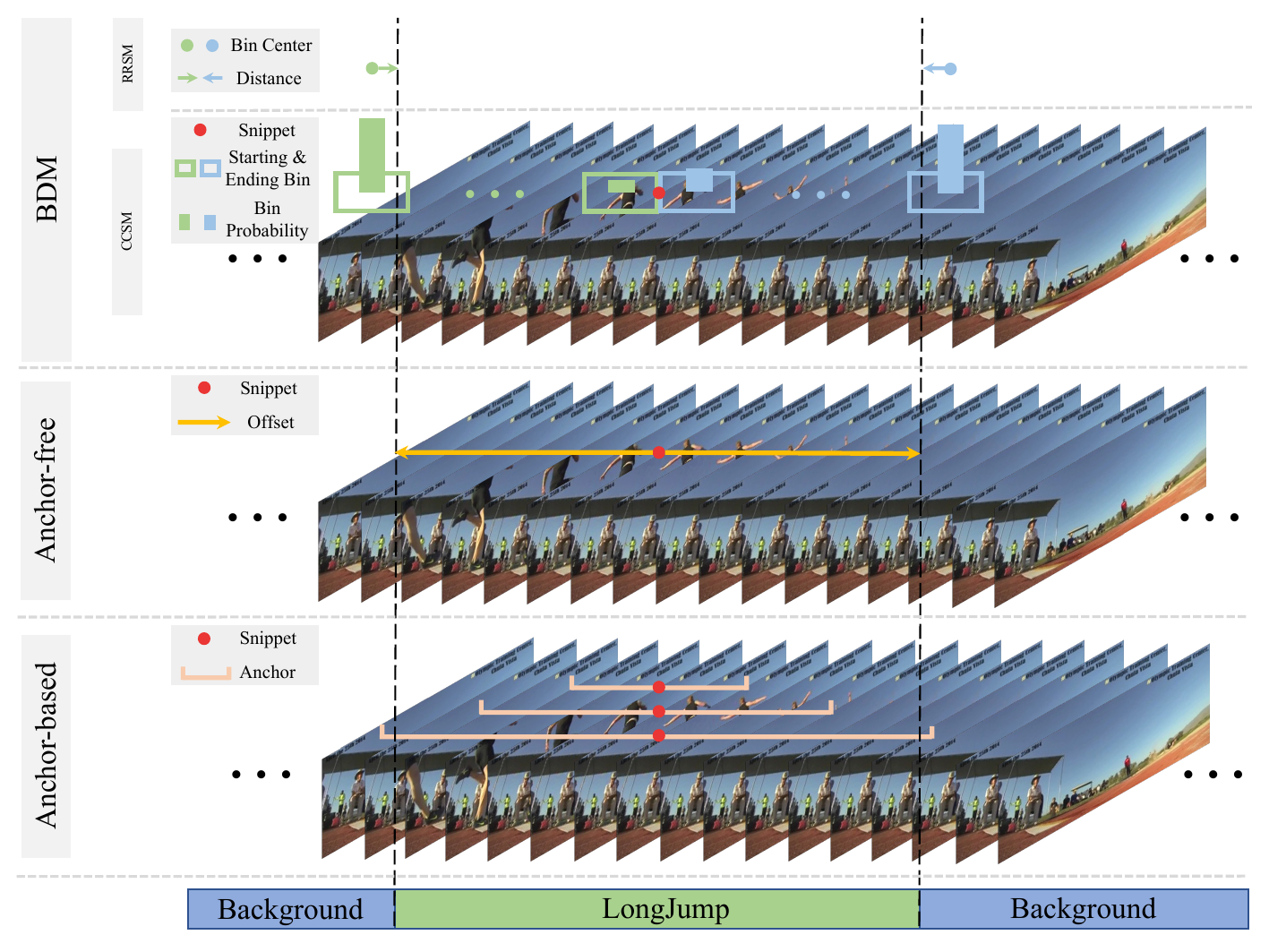}
\end{center}
   \caption{Comparison of boundary discretization, anchor-based, and anchor-free methods for representing action boundaries.}
\label{fig:BDModule_final}
\end{figure}

Despite achieving remarkable detection performance, mixed methods still suffer from two crucial issues: (1) Brute-force merging and handcrafted anchor design affect the performance and practical application of the mixed methods. Existing mixed methods predict action instances with anchor-based and anchor-free branches, respectively, and directly merge these results through post-processing to obtain mixed results. This merging method, through post-processing, increases computational complexity while reducing inference speed. Meanwhile, the anchor-based branch requires carefully handcrafted anchors to achieve optimal performance on different datasets, which could be more conducive to the practical application of TAD. (2) The detection results of existing TAD methods contain a large number of false positive category predictions, which further impact the detection performance. Predicting action categories for each snippet of the entire video in the anchor-free branch creates a large number of false positives due to the impact of non-discriminative snippets. As shown in Table \ref{tab:a2net_remove_fp}, when filtering out false positives from both the mixed method A2Net \cite{a2net} and the anchor-free method ActionFormer \cite{actionformer}, their performance is greatly improved at all tIoU thresholds.

To address the abovementioned issues, we propose a novel framework called BDRC-Net, consisting of a boundary discretization module (BDM) and a reliable classification module (RCM).

(1) As shown in Fig.~\ref{fig:BDModule_final}, BDM represents action boundaries using two sub-modules: coarse classification and refined regression. The coarse classification sub-module (CCSM) discretizes the boundary offsets corresponding to each snippet into multiple bins and predicts which bin the starting and ending boundaries belong to. As each discretized bin encompasses multiple positions, we opt to utilize the central position of each bin as its corresponding location prediction to generate coarse results. Then, the refined regression sub-module (RRSM) enhances the CCSM by regressing the distance between the coarse results and the actual action boundaries. In BDM, the discretized bins can be regarded as pseudo anchors, which preserve the benefit of anchor-based methods in terms of stability. At the same time, the RRSM improves the flexibility of boundary predictions by distance regression. The two sub-modules are elegantly combined by sharing features, avoiding the reliance on carefully designed anchors and addressing the highly computational cost issue of existing mixed methods.

\begin{figure}
\begin{center}
    \includegraphics[width=0.95\linewidth]{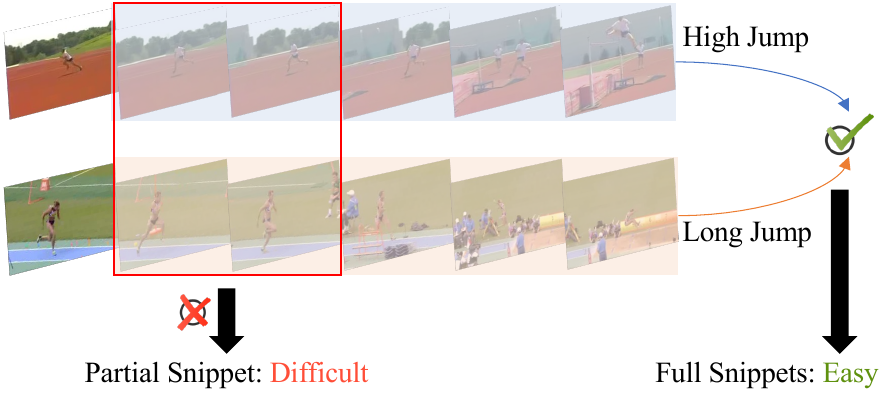}
\end{center}
   \caption{Example of non-discriminative snippets. It is impossible to accurately classify them with only partial snippets from two action categories, namely \textit{HighJump} and \textit{LongJump}. However, if the context of all snippets can be fully utilized, these actions can be easily distinguished.}
\label{fig:non-discreminative}
\end{figure}

(2) Only discriminative snippets are easily recognized in a complete action, while non-discriminative snippets are similar to the background or other action categories. In Fig.~\ref{fig:non-discreminative}, we illustrate the non-discriminative snippets among action instances of different action categories. These non-discriminative snippets have similar motion information, making distinguishing the actual action categories difficult. Anchor-free methods classify the action categories to which each snippet belongs in the entire video, and non-discriminative snippets are prone to classification errors, resulting in false positives. To this end, RCM predicts reliable categories to filter out false positives. Following the design of weakly supervised temporal action detection \cite{untrimmednets, he2022asm, zhao2023novel, lprfzy}, we employ Multiple Instance Learning (MIL) to predict reliable video-level action categories. Specifically, for each action category, RCM selects highly scored discriminative snippets and aggregates their prediction probabilities to obtain reliable video-level probability. These video-level probabilities are used to filter snippet-level predictions.

Extensive experiments on THUMOS'14 and ActivityNet-1.3 demonstrate that our method surpasses the state-of-the-art. Furthermore, the ablation studies validate the effectiveness of each component in BDRC-Net.

In summary, the contributions of this paper are as follows:
\begin{itemize}
\item{We propose an elegant mixed-method boundary discretization module (BDM). BDM predicts action boundaries through a coarse classification sub-module (CCSM) and a refined regression sub-module (RRSM). CCSM discretizes boundary offsets into multiple bins and predicts which bin the action boundaries belong to, resulting in coarse results. RRSM refines the coarse results by regressing the distance between the coarse results and the actual action boundaries.}

\item{We propose a reliable classification module (RCM) for predicting reliable action categories. RCM predicts the action categories occurring in the entire video by aggregating highly discriminative snippets, making its results more reliable than predicting the categories for each snippet individually. It is used to filter out false positives in snippet-level predictions.}

\item{Extensive experimental results demonstrate that our method achieves competitive performance on multiple benchmarks.}

\end{itemize}

\section{Related work}\label{related_work}

\begin{figure*}
\begin{center}
    \includegraphics[width=1.0\linewidth]{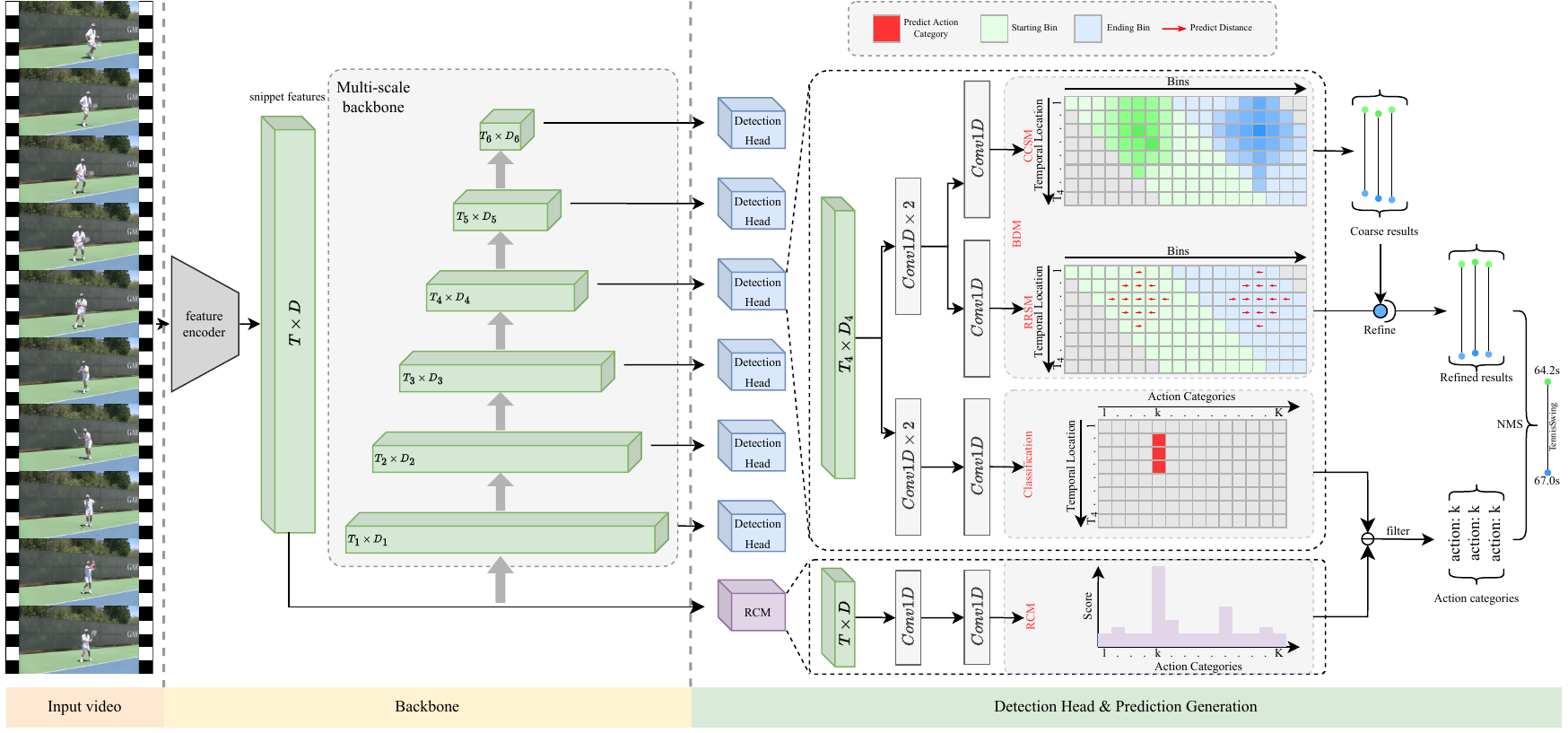}
\end{center}
   \caption{\textbf{Architecture Overview.} Given an untrimmed video, the feature encoder extracts snippet features. Then, the multi-scale backbone (MSB) is used to extract multi-scale spatio-temporal features. Finally, the BDM in the detection head predicts action boundaries on each snippet, and the classification module is used to predict the action categories for each snippet. Specifically, our RCM predicts the video-level action categories based on snippet features, which are used to filter out false positives in the classification module. In the output module of CCSM, \textcolor{green}{green} and \textcolor{blue}{blue} respectively represent the regions of the start and end bins, with \textcolor{gray}{gray} indicating invalid bins. The darker the \textcolor{green}{green} or \textcolor{blue}{blue}, the higher the probability that the corresponding bin is a start or end bin. In the output module of RRSM, the meanings of \textcolor{green}{green} and \textcolor{blue}{blue} are the same as in CCSM. \textcolor{red}{Red} arrows indicate the predicted offset direction by RRSM, with the offset distance prediction omitted for simplicity. In the output of the classification module, \textcolor{red}{red} indicates that the corresponding bin is predicted as the corresponding action category. In the output module of RCM, the longer the \textcolor{lightpurple}{light purple} length, the higher the probability that the input video contains the corresponding action category. BDRC-Net obtains the coarse prediction results through the probabilities predicted by CCSM and refines these results using the offset direction and offset distance predicted by RRSM. The action category of each prediction result is determined jointly by the classification module and the RCM module, with the RCM module primarily used to filter out potential false positive predictions from the classification module.}
\label{fig:summary_model}
\end{figure*}

In this section, we review the previous works related to our paper, which are divided into three parts: (1) anchor-based TAD, (2) anchor-free TAD, (3) query-based TAD, and (4) mixed TAD.

\subsection{Anchor-based TAD}
Anchor-based methods \cite{ssad, decoupssad, rc3d, talnet, gtad, pbrnet, tbos, tallformer} classify each pre-defined anchor obtained from the distribution of action instances in the dataset while also regressing their action boundaries; it can be classified into \textit{one-stage methods} \cite{ssad, decoupssad} and \textit{two-stage methods} \cite{rc3d, gtad, talnet, tbos, tallformer, pbrnet}. One-stage methods predict boundaries for all anchors. SSAD \cite{ssad}, an early one-stage TAD method, uses multi-scale features to predict each anchor's category, overlap, and regression offset. Later, Decouple-SSAD \cite{decoupssad} reduces the mutual influence between the classification and regression branches during multi-task training by decoupling them. Two-stage methods generate action proposals using pre-defined anchors and subsequently classify and perform regression on these proposals. R-C3D \cite{rc3d} was the first to propose an end-to-end trained two-stage method, and its architecture served as the main inspiration for Faster R-CNN \cite{fasterrcnn}. TBOS \cite{tbos} enhances the feature representation for the TAD task by utilizing multi-task learning. More recently TallFormer \cite{tallformer} uses a long-term memory mechanism to capture video information and enable end-to-end training. Anchor-based methods usually set multiple anchors at each position, which implements a dense coverage for action instances, and enhances the stability of the predictions.

\subsection{Anchor-free TAD}
The anchor-based methods depend on the pre-definition of anchors for various datasets, which restricts their flexibility in terms of detection and application. Anchor-free methods \cite{ssn, bsn, bmn, afsd, brem, actionformer, liu2021centerness, lin2019joint, eun2019srg, xu2019cascaded}, on the other hand, do not rely on pre-defined anchors, resulting in more flexible and versatile predictions. SSN \cite{ssn} first introduces the concept of actionness, which uses actionness to generate proposals and then performs two-stage predictions. RAM \cite{ram} introduces relation attention to exploiting informative relations among proposals, further improving the performance of SSN. Subsequently, BSN \cite{bsn} divides action boundaries into starting and ending intervals and performs localization by predicting boundary probabilities. AFSD \cite{afsd} designs a saliency-based refinement module to refine the boundaries of the anchor-free method. Recently, ActionFormer \cite{actionformer} utilizes local self-attention to extract features from videos and introduces a multi-scale transformer block to predict action instances. EAC \cite{react} explicitly modeling action centers to reduce unreliable action detection results caused by ambiguous action boundaries.

\subsection{Query-based TAD}
Query-based TAD introduces a set prediction mechanism to reduce post-processing operations in traditional detectors, ideally avoiding using NMS~\cite{nms} during post-processing. However, most query-based methods still require NMS. RTD-Net~\cite{rtdnet} breaks the one-to-one assignment of the set prediction mechanism, introducing one-to-many matching to alleviate the slow convergence problem of the transformer. ReAct~\cite{react} improves detection performance by modifying the decoder's attention to relational attention but struggles to capture the complete context of queries. TadTR~\cite{tadtr} employs cross-window fusion and uses NMS to remove redundant proposals. Recent research has begun to focus on how to completely avoid using NMS. TE-TAD~\cite{tetad} integrates time-aligned coordinate expression to fully preserve the one-to-one matching paradigm fully, achieving complete end-to-end modeling.

\subsection{Mixed TAD}
Mixed TAD attempts to merge the advantages of the two methods above to achieve more stable and precise detection results with better boundaries. PCAD \cite{pcad} explores enhancing the prediction flexibility by directly merging the boundary probabilities in a two-stage anchor-based method. MGG \cite{mgg} also explores the fusion of anchor-based and boundary probabilities. Unlike PCAD, MGG utilizes the starting and ending probabilities to refine the boundaries of the anchor-based. Recently, A2Net \cite{a2net} conducted a comprehensive comparison between the anchor-based and anchor-free methods, verifying their complementary. When inferencing, A2Net combines the results from the two methods and employs non-maximum suppression \cite{nms} to suppress redundant predictions.

The existing mixed methods violently merge anchor-based and anchor-free methods and depend on handcrafted anchors design. It limits the mixed method's performance and affects the practical application of TAD. Moreover, the false positives in snippet-level action category predictions in the anchor-free branch further reduced the performance of the mixed methods. Our BDRC-Net uses a more elegant merging strategy to address the dependency on handcrafted anchors in existing mixed methods while improving detection performance. At the same time, BDRC-Net further enhances the accuracy of action predictions by predicting reliable action categories.

\section{Method}\label{sec:method}

This section elaborates on the proposed framework. As shown in Fig.~\ref{fig:summary_model}, given an untrimmed video, we extract snippet features through the feature encoder. Moreover, a multi-scale backbone is used to obtain multi-scale features. Next, in the detection head, BDM and the classification module predict each snippet's action boundaries and categories based on multi-scale features. In contrast, RCM directly predicts the action categories contained in the entire video based on the snippet features. Finally, the detection results are obtained by post-processing the results of RCM and the detection head.

\subsection{Problem Definition}\label{subsec:problem_definition}

Given an untrimmed video $V$, we segment the video sequence into $T$ non-overlapping snippets to reduce computational complexity. Specifically, each snippet consists of a small number of consecutive frames. For the $t$-th snippet, we obtain a $D$-dimensional spatio-temporal feature representation $\mathbf{F}_v(t)$ through a pre-trained 3D convolutional network. Collecting the features of all snippets together, we denote it as $\mathbf{F}_v \in \mathbb{R}^{T \times D}$. The annotations of all action instances in $V$ are denoted as $\Psi^g = \{\psi_n^g \mid \psi_n^g = (y_{s,n}^g, y_{e,n}^g, y_{c,n}^g)\}_{n=1}^{N_g}$, where $N_g$ is the number of action instances, and $y_{s,n}^g$, $y_{e,n}^g$, $y_{c,n}^g$ represent the start time, end time, and action label of the $n$-th action instance, respectively. The goal of TAD is to use $\mathbf{F}_v$ to locate starting and ending times of action instances in the untrimmed video and predict their action categories.

\subsection{Multi-Scale Backbone (MSB)}\label{subsec:backbone}

In this subsection, we briefly describe the method for obtaining and representing multi-scale features in a general multi-scale backbone (MSB). An MSB usually consists of multiple base blocks, wherein the first layer of each base block, the video features are downsampled by a factor of $2$ in the temporal dimension. Then some convolutional layers or transformer layers are used to model the spatio-temporal context. It is worth noting that BDRC-Net builds upon the existing MSB, so here we briefly introduce the general workflow of MSB without delving into its specific implementation. The modeled features are used to input the detection head and the next layer of the base block. Therefore, for an $L$-layer MSB, we obtain a set of features: $\{\mathbf{F}_l \mid \mathbf{F}_l \in \mathbb{R}^{T_l \times D_l}\}_{l=1}^{L}$, where $\mathbf{F}_l$ denotes the video features output from the $l$-th layer, and $T_l$, $D_l$ represent the temporal length and dimension of the features at that layer. Specifically, each base block in MSB downsamples the feature's length but keeps the feature's dimension unchanged, so all feature dimensions are the same in MSB. Since the detection head shares weights on the features of each layer in MSB, we use $\mathbf{F}_{ms} \in \mathbb{R}^{T_{ms} \times D_{ms}}$ to represent multi-scale features for clarity in the following parts. Here, $\mathbf{F}_{ms}$ is the concatenation of $\{\mathbf{F}_l\}_{l=1}^{L}$ along the temporal dimension, $T_{ms}=\sum_{l=1}^LT_l$, and $D_{ms} = D_l$. Our proposed methods, BDM and RCM, are suitable for any MSB, and we verify their robustness on different MSB in the experimental section.

\subsection{Boundary Discretization Module (BDM)}\label{subsec:bdm}

As described in Sec.~\ref{sec:introduction}, although mixed methods can utilize the advantages of anchor-based and anchor-free methods, they still suffer from the issues of brute-force merging, handcrafted anchors design, and excessive false positives in action category predictions. To address the issues of brute-force merging and handcrafted anchors design, we propose BDM, which merges anchor-based and anchor-free methods more elegantly. In BDM, we predict action boundaries as a Coarse-Classification Sub-Module (CCSM) and a Refinement Regression Sub-Module (RRSM). CCSM discretizes the distance of action boundaries into multiple bins and predicts which bin the action boundaries belong to. Since each bin represents a small continuous time period, we use the center position of the bin as the predicted coarse boundary. Although this method enhances the stability of boundary predictions, it sacrifices some flexibility. Therefore, we subsequently propose that RRSM refine the coarse boundaries by regressing the distance between the predicted coarse and actual action boundaries.

\begin{figure}
\begin{center}
    \includegraphics[width=1.0\linewidth]{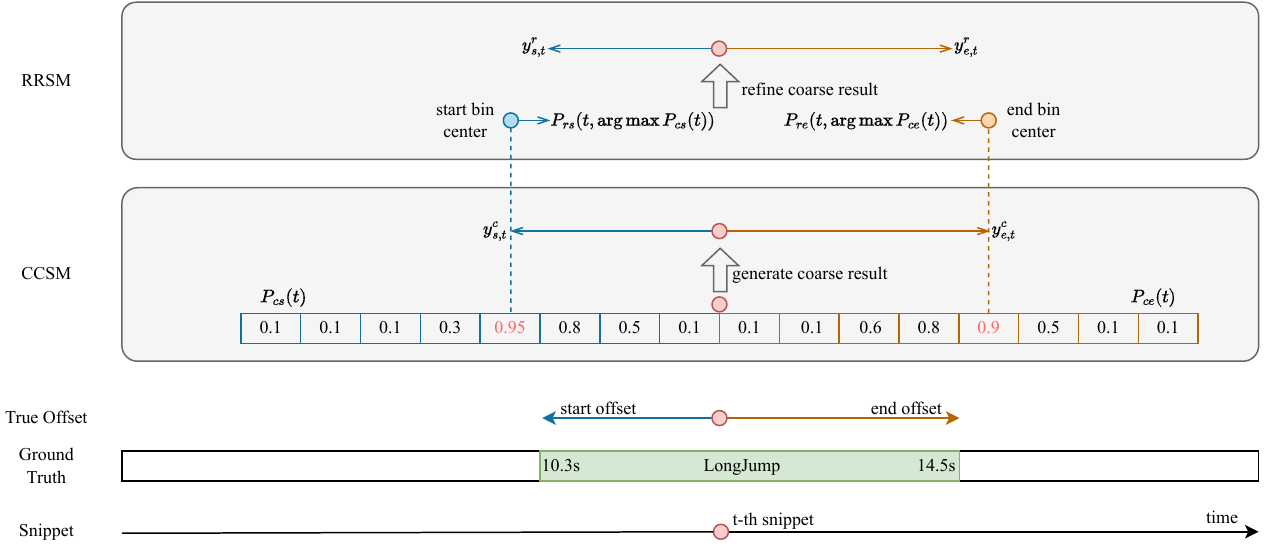}
\end{center}
   \caption{BDM. For the $t$-th snippet, CCSM selects the center position of the bin with the highest predicted probability as a coarse prediction of the action boundary. Subsequently, RRSM regresses the distance between the coarse prediction and the actual boundary to obtain refined boundary prediction.}
\label{fig:BDM_detail}
\end{figure}

\textbf{CCSM.} The purpose of CCSM is to predict the coarse action boundaries. As shown in Fig.~\ref{fig:BDM_detail}, CCSM lays out $W$ bins in both starting and ending directions for the $t$-th snippet and predicts which bin the action boundaries belongs to. Given $\mathbf{F}_{ms}$, mathematically, we have $(P_{cs}, P_{ce}) = \Phi_{ccsm}(\mathbf{F}_{ms})$, where $\Phi_{ccsm}$ represents the CCSM, and $P_{cs} \in \mathbb{R}^{T_{ms} \times W}$, $P_{ce} \in \mathbb{R}^{T_{ms} \times W}$ represents the predictions of starting and ending position, respectively. For the $t$-th snippet, the predicted starting and ending positions are denoted as $P_{cs}(t)$ and $P_{ce}(t)$. As shown in Fig.~\ref{fig:BDM_detail}, we use the center position of the bin with the maximum probability in $P_{cs}(t)$ and $P_{ce}(t)$ as the coarse prediction. Denote the coverage length of each bin as $b$, we can obtain the coarse prediction $y^c_t = (y^c_{s,t}, y^c_{e,t})$ at the $t$-th snippet with the following equations:
\begin{equation}
\begin{cases}
y^c_{s,t}=(t-(\arg\max P_{cs}(t) \times b + b/2)) \times S(t) \\ 
y^c_{e,t}=(t+(\arg\max P_{es}(t) \times b + b/2)) \times S(t)
\end{cases}
\end{equation}
Where $S(t)$ is the scaling factor of $t$-th snippet, if the snippet belongs to the $l$-th level of the MSB, then $S(t) = 2^{l-1}$.

\textbf{RRSM.} CCSM representing the center position of each bin as the boundary results in too coarse predictions, making it difficult to locate action boundaries precisely. Therefore, to further refine the coarse boundaries, RRSM is used to regress the distance between the coarse and actual action boundaries. Given $\mathbf{F}_{ms}$, we have $(P_{rs}, P_{re}) = \Phi_{rrsm}(\mathbf{F}_{ms})$, where $\Phi_{rrsm}$ represents the RRSM, $P_{rs} \in \mathbb{R}^{T_{ms} \times W}$, $P_{re} \in \mathbb{R}^{T_{ms} \times W}$ represents the regression predictions of starting and ending distance, respectively. For the $t$-th snippet, the regression predictions for starting and ending distances corresponding to each bin are denoted as $P_{rs}(t)$ and $P_{re}(t)$, respectively. As shown in Fig.~\ref{fig:BDM_detail}, after obtaining the coarse result through CCSM on the $t$-th snippet, the corresponding regression predictions on starting and ending bins are used to refine the coarse result. Finally, we obtain the refined result $y^r_t = (y^r_{s,t}, y^r_{e,t})$ on the $t$-th snippet with the following equations:
\begin{equation}
\begin{cases}
y^r_{s,t}=y^c_{s,t}-P_{rs}(t,\arg\max P_{cs}(t)) \\ 
y^r_{e,t}=y^c_{e,t}+P_{re}(t,\arg\max P_{es}(t))
\end{cases}
\end{equation}

\subsection{Reliable Classification Module (RCM)}\label{subsec:rcm}

The purpose of RCM is to predict the reliable action categories that occur throughout the entire video to filter out false positives caused by traditional snippet-level action classification. In this sub-section, we first introduce the typical snippet-level action classification and then provide a detailed description of the workflow of RCM.

\textbf{Snippet-level action classification.} Existing one-stage anchor-free methods use snippet-level action classification to predict the action categories to which each input snippet belongs. Let the number of action categories be $K$, mathematically we have: $P_{sc}=\Phi_{sc}(\mathbf{F}_{ms})$. Here, $\Phi_{sc}$ is the snippet-level classification network, and $P_{sc} \in \mathbb{R}^{T_{ms} \times K}$ is the action classification probabilities. Furthermore, we denote $t$-th snippet's corresponding prediction probabilities as $P_{sc}(t)$. If the prediction probability $P_{sc}(t,k)$ of the $k$-th action category on the $t$-th snippet is greater than $\lambda_{cls}$, this indicates that $k$-th action occurs on that snippet. It should be noted that there may be multiple action categories with prediction probabilities greater than $\lambda_{cls}$ on the given snippet, resulting in multiple predicted action categories for that snippet where $\lambda_{cls}$ is a hyper-parameter of predict action categories. While improving the recall rate of action instances, this method also increases the false positives in the predicted results.

\textbf{RCM.} To reduce false positives caused by snippet-level action classification, RCM is proposed. RCM predicts reliable video-level action categories by aggregating the prediction results of discriminative snippets. Unlike BDM and snippet-level classification, RCM directly uses snippet features $\mathbf{F}_v$ as input instead of using multi-scale features $\mathbf{F}_{ms}$. In snippet-level classification, $\mathbf{F}_{ms}$ is used to classify each snippet's actions. However, the snippet-level classification results on non-discriminative segments, as shown in Fig.~\ref{fig:non-discreminative}, are untrustworthy. Even if these results can help us find the location of action instances with high activation values, the predicted action categories may be wrong. In contrast, the video-level action predictions of RCM do not need to consider the location of action instances. It only needs to aggregate the predictions of some highly discriminative snippets to obtain reliable action categories for the entire video. Therefore, using $\mathbf{F}_v$ in RCM can avoid the mutual disturbance caused by inconsistencies with the snippet-level classification target when using $\mathbf{F}_{ms}$.

Given the snippet feature $\mathbf{F}_v$, RCM first predicts the action categories for each snippet. Mathematically, we have: $P_{rc}=\Phi_{rcm}(\mathbf{F}_v)$. Here, $\Phi_{rcm}$ is the RCM network, and $P_{rc} \in \mathbb{R}^{T \times K}$ is the predicted action probabilities for each snippet. Specifically, to obtain reliable video-level categories, RCM selects the top-$N_v$ snippets with the highest confidence score as highly discriminative snippets. Moreover, we use $P_{rc}$ as the confidence score. In contrast, the remaining snippets are denoted as non-discriminative snippets. In RCM, only the highly discriminative snippets are used to predict the probabilities $P_{vl} \in \mathbb{R}^K$ of video-level categories. Especially, $P_{vl}(k)$ is the probability of the $k$-th action occurring in the video, which is calculated by the following equation:
\begin{equation}
    P_{vl}(k) = \max_{\substack{\mathbf{M} \subset \mathcal{P}_{rc}(:, k) \\ |\mathcal{M}| = N_v}} \frac{1}{N_v} \sum_{r=1}^{N_v}\mathbf{M}(r)
\end{equation}
Where, $\mathbf{M} \in \mathbb{R}^{N_v}$ is the set of probability values for the top-$N_v$ snippets with the highest predicted probabilities for the $k$-th action category. Finally, we obtain a reliable set of action categories $y^{vid}=\{k \mid P_{vl}(k) > \lambda_{vid}\}$, where $\lambda_{vid}$ is a hyper-parameter used to obtain action categories based on prediction probabilities.

\section{Training and inference}\label{sec:train_and_inference}

\begin{figure}
\begin{center}
    \includegraphics[width=1.0\linewidth]{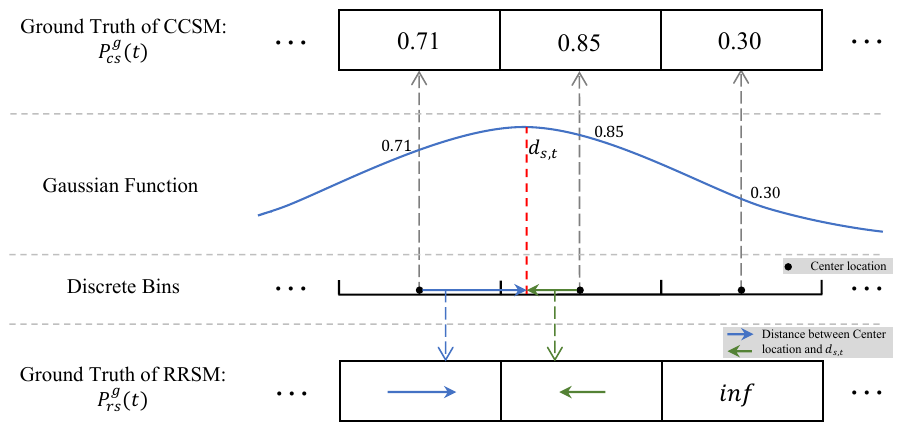}
\end{center}
   \caption{Label assignment in BDM. For the starting bin on the $t$-th snippet, firstly, the distance $d_{s,t}$ between snippet $t$ and the starting boundary is computed. Then, a Gaussian function with mean and variance as $d_{s,t}$ and $\sigma$ is used to assign labels in CCSM. Specifically, the label for each bin in CCSM is the value of its central position on the Gaussian function. We further assign labels for RRSM, where the label for each bin on RRSM is the distance between $d_{s,t}$ and its center position.}
\label{fig:gt_details_v2}
\end{figure}

In this section, we provide detailed explanations of the training and inference processes for the proposed method.

\subsection{Training}\label{subsec:training}

In the training section, we first introduce label assignment and then compute the loss function based on the results of label assignment and the model prediction in Sec.~\ref{sec:method}.

\textbf{Label Assignment.} Given the annotation set $\Psi^g = \{\psi_n^g \mid \psi_n^g = (y_{s,n}^g, y_{e,n}^g, y_{c,n}^g)\}_{n=1}^{N_g}$ for video $V$, we generate corresponding ground truths for CCSM, RRSM, snippet-level classification and RCM. The ground truth of the starting and ending bins in CCSM is denoted as $P^g_{cs} \in \mathbb{R}^{T_{ms} \times W}$ and $P^g_{ce} \in \mathbb{R}^{T_{ms} \times W}$, while the ground truth of the starting and ending bins in RRSM is denoted as $P^g_{rs} \in \mathbb{R}^{T_{ms} \times W}$ and $P^g_{re} \in \mathbb{R}^{T_{ms} \times W}$. For snippet-level classification and RCM, the ground truth is denoted as $P^g_{sc} \in \mathbb{R}^{T_{ms} \times K}$ and $P^g_{vl} \in \mathbb{R}^{K}$, respectively. In particular, the initial values of $P^g_{cs}$, $P^g_{ce}$, $P^g_{sc}$, and $P^g_{vl}$ are set to $0$, while the initial values of $P^g_{rs}$ and $P^g_{re}$ are set to $\inf$.

As shown in Fig.~\ref{fig:gt_details_v2}, if the $t$-th snippet belongs to the annotation of the $n$-th action instance, i.e., $y_{s,n}^g \leq t \leq y_{e,n}^g$, then the ground truth for snippet-level classification corresponding to this snippet is $P^g_{sc}(t,y_{c,n}^g)=1$. Furthermore, the distance between this snippet and the starting position of the $n$-th annotation is denoted as $d_{s,t}$, where $d_{s,t}= t - y_{s,n}^g / S(t)$. We assign ground truth to each starting bin $P^g_{cs}(t, :)$ of CCSM on the $t$-th snippet by convolving a Dirac delta function centered at $d_{s,t}$ with a Gaussian of fixed variance $\sigma$. Specifically, the ground truth of the $w$-th bin in $P^g_{cs}(t)$ denote as $P^g_{cs}(t,w)$, mathematically, we have:
\begin{equation}
    P^g_{cs}(t,w)=\frac{1}{\sigma\sqrt{2\pi}}e^{-\frac{1}{2}(\frac{x_w - d_{s,t}}{\sigma})^2}
\end{equation}
Where $x_w$ is the center position of the $w$-th starting bin on the $t$-th snippet, which is $x_w=w \times b+b/2$. We further assign ground truth to each starting bin $P^g_{rs}(t, :)$ of RRSM on the $t$-th snippet. Specifically, for the $w$-th starting bin on this snippet, if $P^g_{cs}(t,w) > \lambda_{rs}$, then it's ground truth $P^g_{rs}(t,w)$ is the distance between its center position and the starting boundary of the $n$-th annotation: $P^g_{rs}(t,w)=(d_{s,t}-x_w) \times S(t)$. Similarly, we assign the corresponding ground truths $P^g_{ce}(t)$ and $P^g_{re}(t)$ of CCSM and RRSM to the ending bins of the $t$-th snippet.

After assigning ground truth to all snippets, we obtain $P^g_{cs}$, $P^g_{ce}$, $P^g_{rs}$, $P^g_{re}$, and $P^g_{sc}$. RCM is used to predict the action categories that occur in the video; the ground truth $P^g_{vl}$ can be obtained by aggregating $P^g_{sc}$. Specifically, $P^g_{vl}(k) = \mathbb{I}[\sum_{t=1}^{T_{ms}}P^g_{sc}(t, k) > 0]$, where $\mathbb{I}[\cdot]$ is the indicator function, and $P^g_{vl}(k)$ is the label of $k$-th action in $P^g_{vl}$.

\textbf{Loss Function.} Given the predictions $P_{cs}$ and $P_{ce}$ of CCSM and their corresponding ground truths $P^g_{cs}$ and $P^g_{ce}$, the loss function $\mathcal{L}_{ccsm}$ of CCSM is calculated by the following equation:
\begin{equation}
\begin{aligned}
    \mathcal{L}_{ccsm} = \frac{1}{\lambda_{norm}}(\sum_{t=1}^{T_{ms}}\sum_{w=1}^{W}(\mathcal{L}_{focal}(P_{cs}(t,w), P^g_{cs}(t,w)) \\
    + \mathcal{L}_{focal}(P_{ce}(t,w), P^g_{ce}(t,w))))
\end{aligned}
\end{equation}
Where $\mathcal{L}_{focal}$ represents the focal loss \cite{focalloss}, and $\lambda_{norm}$ is the coefficient used to balance the loss.

For RRSM, we calculate the smooth $L_1$ loss $\mathcal{L}_{sl}$ on all positive bins whose ground truths is non-$\inf$, resulting in $\mathcal{L}_{rrsm}$:
\begin{equation}
\begin{aligned}
    \mathcal{L}_{rrsm} = \frac{1}{\lambda_{norm}}(\sum_{t=1}^{T_{ms}}\sum_{w=1}^{W}(& \mathbb{I}[P^g_{rs}(t,w) \neq \inf] \\
    &\times \mathcal{L}_{sl}(P_{rs}(t,w), P^g_{rs}(t,w)) \\
    &+ \mathbb{I}[P^g_{re}(t,w) \neq \inf] \\
    &\times \mathcal{L}_{sl}(P_{re}(t,w), P^g_{re}(t,w))))
\end{aligned}
\end{equation}
Where $\mathbb{I}[\cdot]$ is the indicator function.

The cross-entropy loss $\mathcal{L}_{ce}$ is used to calculate the snippet-level action classification loss $\mathcal{L}_{cls}$ and the video-level classification loss $\mathcal{L}_{rcm}$ of RCM:

\begin{equation}
\begin{aligned}
    \mathcal{L}_{cls} &= \frac{1}{\lambda_{norm}}\sum_{t=1}^{T_ms}\sum_{k=1}^{K} \mathcal{L}_{ce}(P_{sc}(t,k), P^g_{sc}(t,k)) \\
    \mathcal{L}_{rcm} &= \frac{1}{\lambda_{norm}}\sum_{k=1}^{K} \mathcal{L}_{ce}(P_{vl}(k), P^g_{vl}(k))
\end{aligned}
\end{equation}

Finally, we calculate the sum of $\mathcal{L}_{ccsm}$, $\mathcal{L}_{rrsm}$, $\mathcal{L}_{cls}$, and $\mathcal{L}_{rcm}$ to obtain the overall loss $\mathcal{L}$ of our model:
\begin{equation}
    \mathcal{L} = \mathcal{L}_{ccsm} + \mathcal{L}_{rrsm} + \mathcal{L}_{cls} + \mathcal{L}_{rcm}
\end{equation}

\subsection{Inference}

Given an untrimmed video, we obtain the video-level action category results in $y^{vid}$ using RCM as described in Sec.~\ref{subsec:rcm}. Firstly, we collect a set of all possible locations that may contain actions:
\begin{equation}
    \Psi^{lo} = \{(t_i,k_i) \mid P_{sc}(t_i,k_i) > \lambda_{cls} \wedge k_i \in y^{vid}\}_{i=1}^{N_p}
\end{equation}
Where $t_i$ and $k_i$ respectively represent the snippet index and action category index corresponding to the $i$-th predicted location, and $N_p$ is the total number of predicted locations. For location $(t_i,k_i)$ in $\Psi^{lo}$, we take the boundary prediction $y^r_{t_i}=(y^r_{s,t_i}, y^r_{e,t_i})$ of the corresponding snippet in Sec.~\ref{subsec:bdm} as the starting and ending boundary. Secondly, we calculate the confidence score for each location, where for the $i$-th prediction, the confidence score $c_i$ is defined as:
\begin{equation}
    c_i = P_{sc}(t_i,k_i) \times \sqrt{\max P_{cs}(t_i) \times \max P_{ce}(t_i)}
\end{equation}
Finally, we obtain a set of predicted action instances: $\Psi=\{\psi_i \mid \psi_i = (y^r_{s,t_i}, y^r_{e,t_i}, k_i, c_i)\}_{i=1}^{N_p}$, where $y^r_{s,t_i}$, $y^r_{e,t_i}$, $k_i$, and $c_i$ respectively represent the starting boundary, ending boundary, action category, and confidence score predicted for the $i$-th action instance. Specially, we use non-maximum suppression (NMS) \cite{nms} to suppress redundant action instances with high overlap.

\section{experiments}\label{sec:experiments}

We conducted extensive experiments on the THUMOS'14 \cite{thumos14}, and ActivityNet-1.3 \cite{activitynet1.3} benchmarks and compared our experimental results with previous works. Additionally, we performed extensive ablation experiments to validate our contributions.

\begin{table*}[t!]
    \caption{Temporal action detection results on THUMOS'14, measured by mAP at different tIoU thresholds. The best and second-best results in each column are marked in \textcolor{red}{red} and \textcolor{blue}{blue}, respectively. Commonly used feature encoders include: C3D \cite{c3d}, P3D \cite{p3d}, I3D \cite{i3d}, R(2+1)D \cite{r21d}, TSN \cite{tsn}, SF \cite{slowfast}, VGG \cite{vgg}, VideoMAE~\cite{tong2022videomae}, VideoMAE V2~\cite{wang2023videomae}, and InternVideoV2~\cite{wang2024internvideo2}. Avg.: average mAP under tIoU thresholds $[0.3:0.1:0.7]$. AB, AF, and Query represent Anchor-Based, Anchor-Free, and Query-Based methods, respectively.}
    \centering
    \resizebox{0.8\linewidth}{!}{\begin{tabular}{c|c|c|c|cccccc}
\hline
 &  &  &  & \multicolumn{6}{c}{\textbf{mAP@tIoU(\%)}} \\ \cline{5-10} 
\multirow{-2}{*}{\textbf{Type}} & \multirow{-2}{*}{\textbf{Model}} & \multirow{-2}{*}{\textbf{Feature}} & \multirow{-2}{*}{\textbf{Venue}} & \textbf{0.3} & \textbf{0.4} & \textbf{0.5} & \textbf{0.6} & \textbf{0.7} & \cellcolor[HTML]{C0C0C0}\textbf{Avg.} \\ \hline \hline
 & R-C3D~\cite{rc3d} & C3D & \textit{ICCV'17} & 44.8 & 35.6 & 28.9 & - & - & \cellcolor[HTML]{C0C0C0}- \\
 & PBR-Net~\cite{pbrnet} & I3D & \textit{AAAI'20} & 58.5 & 54.6 & 51.3 & 41.8 & 29.5 & \cellcolor[HTML]{C0C0C0}47.1 \\
 & Gemini~\cite{gemini} & Gemini & TMM'20 & 56.7 & 50.6 & 42.6 & 32.5 & 21.4 & \cellcolor[HTML]{C0C0C0}40.8 \\
 & STAN~\cite{stan} & TSN+VGG & TMM'21 & 67.5 & 61.0 & 51.7 & - & - & \cellcolor[HTML]{C0C0C0}- \\
 & MUSES~\cite{muses} & I3D & \textit{CVPR'21} & 68.9 & 64.0 & 56.9 & 46.3 & 31.0 & \cellcolor[HTML]{C0C0C0}53.4 \\
 & VSGN~\cite{vsgn} & I3D & \textit{ICCV'21} & 66.7 & 60.4 & 52.4 & 41.0 & 30.4 & \cellcolor[HTML]{C0C0C0}50.1 \\
\multirow{-7}{*}{AB} & SAC~\cite{yang2022structured} & I3D & \textit{TIP'22} & 69.3 & 64.8 & 57.6 & 47.0 & 31.5 & \cellcolor[HTML]{C0C0C0}54.0 \\ \hline \hline
 & AFSD~\cite{afsd} & I3D & \textit{CVPR'21} & 67.3 & 62.4 & 55.5 & 43.7 & 31.1 & \cellcolor[HTML]{C0C0C0}52.0 \\
 & E2E-TAD~\cite{e2etad} & SF R50 & \textit{CVPR'22} & 69.4 & 64.3 & 56.0 & 46.4 & 34.9 & \cellcolor[HTML]{C0C0C0}54.2 \\
 & TAGS~\cite{tags} & I3D & \textit{ECCV'22} & 68.6 & 63.8 & 57.0 & 46.3 & 31.8 & \cellcolor[HTML]{C0C0C0}52.8 \\
 & ActionFormer~\cite{actionformer} & I3D & \textit{ECCV'22} & 82.1 & 77.8 & 71.0 & 59.4 & 43.9 & \cellcolor[HTML]{C0C0C0}66.8 \\
 & TAGS+GAP~\cite{gap} & I3D & \textit{CVPR'23} & 69.1 & - & 57.4 & - & 32.0 & \cellcolor[HTML]{C0C0C0}53.0 \\
 & MENet~\cite{menet} & R(2+1)D & \textit{ICCV'23} & 70.7 & 65.3 & 58.8 & 49.1 & 34.0 & \cellcolor[HTML]{C0C0C0}55.6 \\
 & ASL~\cite{asliccv23} & I3D & \textit{ICCV'23} & 83.1 & 79.0 & 71.7 & 59.7 & 45.8 & \cellcolor[HTML]{C0C0C0}67.9 \\
 & TriDet~\cite{tridet} & I3D & \textit{CVPR'23} & 83.6 & 80.1 & 72.9 & 62.4 & 47.4 & \cellcolor[HTML]{C0C0C0}69.3 \\
 & AdaTAD~\cite{adatad} & VideoMAE-B & \textit{CVPR'24} & 86.0 & 81.9 & 75.0 & 63.3 & 49.7 & \cellcolor[HTML]{C0C0C0}71.1 \\
 & ViT-TAD~\cite{vittad} & VideoMAE-V2 & CVPR'24 & 85.1 & 80.9 & 74.2 & 61.8 & 45.4 & \cellcolor[HTML]{C0C0C0}69.5 \\
 & ADSFormer~\cite{adsformer} & I3D & TMM'24 & 84.4 & 80.0 & 73.1 & 62.9 & 46.9 & \cellcolor[HTML]{C0C0C0}69.5 \\
 & ADSFormer~\cite{adsformer} & VideoMAE-V2 & TMM'24 & 85.3 & 80.8 & 73.9 & 64.0 & 49.8 & \cellcolor[HTML]{C0C0C0}70.8 \\
\multirow{-13}{*}{AF} & ActionMamba~\cite{actionmamba} & InternVideoV2-6B & arXiv'24 & \textcolor{blue}{86.9} & \textcolor{blue}{83.1} & \textcolor{blue}{76.9} & \textcolor{blue}{65.9} & \textcolor{blue}{50.8} & \cellcolor[HTML]{C0C0C0}\textcolor{blue}{72.7} \\ \hline \hline
 & TadTR~\cite{tadtr} & I3D & TIP'22 & 74.8 & 69.1 & 60.1 & 46.6 & 32.8 & \cellcolor[HTML]{C0C0C0}56.7 \\
 & RTD-Net~\cite{rtdnet} & I3D & \textit{ICCV'21} & 68.3 & 62.3 & 51.9 & 38.8 & 23.7 & \cellcolor[HTML]{C0C0C0}49.0 \\
 & ReAct~\cite{react} & TSN & \textit{ECCV'22} & 69.2 & 65.0 & 57.1 & 47.8 & 35.6 & \cellcolor[HTML]{C0C0C0}55.0 \\
 & Self\_DETR~\cite{selfdetr} & I3D & ICCV'23 & 74.6 & 69.5 & 60.0 & 47.6 & 31.8 & \cellcolor[HTML]{C0C0C0}56.7 \\
 & DualDETR~\cite{dualdetr} & I3D & CVPR'24 & 82.9 & 78.0 & 70.4 & 58.5 & 44.4 & \cellcolor[HTML]{C0C0C0}66.8 \\
\multirow{-6}{*}{Query} & TE-TAD~\cite{tetad} & I3D & CVPR'24 & 83.3 & 78.4 & 71.3 & 60.7 & 45.6 & \cellcolor[HTML]{C0C0C0}67.9 \\ \hline \hline
 & PCAD~\cite{pcad} & C3D & \textit{TOMM'20} & 52.8 & 47.1 & 39.2 & - & - & \cellcolor[HTML]{C0C0C0}- \\
 & MGG~\cite{mgg} & TSN & \textit{CVPR'19} & 53.9 & 46.8 & 37.4 & 29.5 & 21.3 & \cellcolor[HTML]{C0C0C0}37.8 \\
 & A2Net~\cite{a2net} & I3D & \textit{TIP'20} & 58.6 & 54.1 & 45.5 & 32.5 & 17.2 & \cellcolor[HTML]{C0C0C0}41.6 \\ \cline{2-10} 
 & A2Net+ours & I3D & \textit{-} & 65.9 & 60.0 & 50.6 & 38.6 & 25.2 & \cellcolor[HTML]{C0C0C0}48.1 \\
\multirow{-5}{*}{Mixed} & TriDet+ours & InternVideoV2-6B & - & \textcolor{red}{88.6} & \textcolor{red}{83.8} & \textcolor{red}{78.0} & \textcolor{red}{66.8} & \textcolor{red}{51.2} & \cellcolor[HTML]{C0C0C0}\textcolor{red}{73.7} \\ \hline
\end{tabular}}
\label{tab:thumos_main}
\end{table*}

\subsection{Datasets}\label{subsec:datasets}

\textbf{THUMOS'14} consists of $200$ validation videos and $213$ test videos with boundary annotations for $20$ action categories. Following the standard setting \cite{actionformer, tags, tallformer, a2net, afsd}, we train our model on the validation videos and evaluate its performance on the test videos.

\textbf{ActivityNet-1.3} consists of $19,994$ videos with action category annotations for 200 classes. It is split into train, validation, and test sets in a $2:1:1$ ratio. We train our model on the train set and evaluate its performance on the validation set.

\textbf{MultiTHUMOS} is an extension of THUMOS'14, with the same number of videos. The number of action categories in MultiTHUMOS has increased to $65$. Each video has an average of $97$ ground-truth instances and $10.5$ action categories.

\subsection{Evaluation Metric}\label{subsec:evaluation_metric}

Mean average precision (mAP) is a commonly used metric for evaluating the performance of TAD models and is measured using various temporal intersection over union (tIoU) thresholds. tIoU is calculated as the intersection over the union between two temporal boundaries. To evaluate the model's performance, mAP computes the mean of average precision scores across all action categories using a given tIoU threshold.

\subsection{Implementation Details}\label{subsec:implementation_details}

For the feature encoder, we divide every 16 frames into one snippet and use a pre-trained two-stream I3D \cite{i3d} network on Kinetics \cite{i3d} to extract snippet features. Our method is suitable for any existing multi-scale backbone (MSB). Specifically, we conduct experiments on MSB implemented in ActionFormer \cite{actionformer}, which has achieved excellent performance, and the MSB of A2Net \cite{a2net}, a state-of-the-art mixed method. Our CCSM, RRSM, and snippet-level classification are built using three temporal convolution layers with consistent parameter settings, including kernel size, stride, padding, and dimension, which are set to $3$, $1$, $1$, and $512$, respectively. Expressly, for the prediction layer, the dimensions of CCSM, RRSM, and snippet-level classification are set to $2 \times W$, $2 \times W$, and $K$, respectively. For RCM, we use two temporal convolution layers with identical parameters for kernel size, stride, and padding, which are set to $3$, $1$, and $1$, respectively. The dimension of the first and second layers are set to $512$ and $K$, respectively. Furthermore, $N_v$ is set as $T_{ms}/8$ to aggregate highly discriminative snippets. All hyper-parameters are determined by empirical grid search. For the bin, we set $W$ and $b$ to $8$ and $0.25$, respectively, and the $\sigma$ used for label assignment of CCSM is set to $\sqrt{0.2}$. For the hyper-parameters $\lambda_{vid}$, $\lambda_{rs}$, and $\lambda_{norm}$, we set them to $0.1$, $0.5$, and $90$, respectively. The hyper-parameter $\lambda_{cls}$ follows the setting used in MSB, and we do not make any specific adjustments. During training, we train for $34$ epochs on the THUMOS'14 dataset with a learning rate of $1e-4$ and batch size of $2$ and decay the learning rate to $1e-5$ at the $30$-th epoch. For the ActivityNet-1.3 dataset, we train for $15$ epochs with a learning rate of $5e-4$ and batch size of $16$. During inference, the NMS threshold is set to $0.2$, and following the standard setting \cite{actionformer, a2net}, we keep up to $200$ predict results per video based on the confidence score.

\subsection{Main Results}\label{subsec:main_results}

\textbf{THUMOS'14.} On the THUMOS'14 dataset, we report the performance of our method and compare the results with previous works in Table~\ref{tab:thumos_main}. It can be observed that our approach with TriDet backbone achieves an average mAP of $73.7\%$, surpassing the previous state-of-the-art method. When employing the A2Net backbone, our approach exhibits performance improvements like our counterparts. It is worth noting that, when using InternVideoV2-6B features, our method significantly outperforms ActionMamba regarding average mAP ($72.7\% \, vs. \, 73.7\%$). This demonstrates the effectiveness of our approach. In the following sections, we provide comparisons of BDRC-Net with existing state-of-the-art methods using different backbones and features to validate our method's robustness further.

\begin{table}[t!]
    \caption{Temporal action detection results on ActivityNet-1.3, measured by mAP at different tIoU thresholds. The best and second-best results in each column are marked in \textcolor{red}{red} and \textcolor{blue}{blue}, respectively. Avg.: average mAP under tIoU thresholds $[0.5:0.05:0.95]$. AB, AF, and Query represent Anchor-Based, Anchor-Free, and Query-Based methods, respectively.}
    \centering
    \resizebox{0.9\linewidth}{!}{\begin{tabular}{c|c|c|cccc}
\hline
 &  &  & \multicolumn{4}{c}{\textbf{mAP@tIoU(\%)}} \\ \cline{4-7} 
\multirow{-2}{*}{\textbf{Type}} & \multirow{-2}{*}{\textbf{Model}} & \multirow{-2}{*}{\textbf{Feature}} & \textbf{0.5} & \textbf{0.75} & \textbf{0.95} & \cellcolor[HTML]{C0C0C0}\textbf{Avg.} \\ \hline \hline
 & PBR-Net~\cite{pbrnet} & I3D & 53.9 & 34.9 & 9.0 & \cellcolor[HTML]{C0C0C0}35.0 \\
 & MUSES~\cite{muses} & I3D & 50.0 & 34.9 & 6.6 & \cellcolor[HTML]{C0C0C0}34.0 \\
\multirow{-3}{*}{AB} & VSGN~\cite{vsgn} & I3D & 52.3 & 36.0 & 8.4 & \cellcolor[HTML]{C0C0C0}35.0 \\ \hline
 & AFSD~\cite{afsd} & I3D & 52.4 & 34.3 & 6.5 & \cellcolor[HTML]{C0C0C0}34.4 \\
 & E2E-TAD~\cite{e2etad} & SF R50 & 50.5 & 36.0 & 10.5 & \cellcolor[HTML]{C0C0C0}35.1 \\
 & TAGS~\cite{tags} & I3D & 56.3 & 36.8 & 9.6 & \cellcolor[HTML]{C0C0C0}36.5 \\
 & ActionFormer\_multi~\cite{actionformer} & I3D & 46.4 & 31.3 & 6.4 & \cellcolor[HTML]{C0C0C0}30.6 \\
 & ActionFormer\_multi~\cite{actionformer} & TSP & 52.1 & 34.9 & 7.2 & \cellcolor[HTML]{C0C0C0}34.2 \\
 & ActionFormer~\cite{actionformer} & I3D & 53.5 & 36.2 & 8.2 & \cellcolor[HTML]{C0C0C0}35.6 \\
 & ActionFormer~\cite{actionformer} & TSP & 54.7 & 37.8 & 8.4 & \cellcolor[HTML]{C0C0C0}36.6 \\
 & MENet~\cite{menet} & TSP & 54.7 & 38.4 & 10.5 & \cellcolor[HTML]{C0C0C0}37.7 \\
 & ASL~\cite{asliccv23} & TSP & 54.1 & 37.4 & 8.0 & \cellcolor[HTML]{C0C0C0}36.2 \\
 & TriDet~\cite{tridet} & TSP & 54.7 & 38.0 & 8.4 & \cellcolor[HTML]{C0C0C0}36.8 \\
 & AdaTAD~\cite{adatad} & VideoMAE-B & 56.8 & 39.4 & 9.7 & \cellcolor[HTML]{C0C0C0}38.4 \\
 & ViT-TAD~\cite{vittad} & VideoMAE-V2 & 55.9 & 38.5 & 8.8 & \cellcolor[HTML]{C0C0C0}37.4 \\
 & ADSFormer~\cite{adsformer} & TSP & 55.3 & 38.4 & 8.4 & \cellcolor[HTML]{C0C0C0}37.1 \\
 & ActionFormer~\cite{actionformer} & InternVideoV2-6B & \textcolor{blue}{61.5} & \textcolor{red}{44.6} & \textcolor{red}{12.7} & \cellcolor[HTML]{C0C0C0}\textcolor{blue}{41.2} \\
 & ActionMamba~\cite{actionmamba} & InternVideoV2-6B & \textcolor{red}{62.4} & \textcolor{blue}{43.5} & 10.2 & \cellcolor[HTML]{C0C0C0}\textcolor{red}{42.0} \\
 & ActionFormer\_multi~\cite{actionformer} & InternVideoV2-6B & 60.5 & 41.5 & 9.3 & \cellcolor[HTML]{C0C0C0}40.4 \\
\multirow{-17}{*}{AF} & TriDet\_multi~\cite{tridet} & InternVideoV2-6B & 58.0 & 38.8 & 9.0 & \cellcolor[HTML]{C0C0C0}38.4 \\ \hline
 & RTD-Net~\cite{rtdnet} & I3D & 47.2 & 30.7 & 8.6 & \cellcolor[HTML]{C0C0C0}30.8 \\
 & Self-DETR~\cite{selfdetr} & I3D & 52.3 & 33.7 & 8.4 & \cellcolor[HTML]{C0C0C0}33.8 \\
 & DualDETR~\cite{dualdetr} & I3D & 52.6 & 32.0 & 7.8 & \cellcolor[HTML]{C0C0C0}34.3 \\
\multirow{-4}{*}{Query} & TE-TAD~\cite{tetad} & TSP & 54.2 & 38.1 & \textcolor{blue}{10.6} & \cellcolor[HTML]{C0C0C0}37.1 \\ \hline \hline
 & A2Net~\cite{a2net} & I3D & 43.6 & 28.7 &  & \cellcolor[HTML]{C0C0C0}27.8 \\ \cline{2-7} 
 & ActionFormer\_multi+ours & I3D & 50.5 & 34.6 &  & \cellcolor[HTML]{C0C0C0}33.7 \\
 & ActionFormer\_multi+ours & TSP & 53.0 & 36.4 & 7.3 & \cellcolor[HTML]{C0C0C0}35.4 \\
 & ActionFormer+ours & I3D & 54.5 & 37.9 &  & \cellcolor[HTML]{C0C0C0}36.8 \\
 & ActionFormer+ours & TSP & 54.5 & 38.2 & 7.4 & \cellcolor[HTML]{C0C0C0}36.9 \\
\multirow{-6}{*}{Mixed} & ActionFormer\_multi+ours & InternVideoV2-6B & 61.4 & 42.1 & 10.2 & \cellcolor[HTML]{C0C0C0}41.1 \\ \hline
\end{tabular}}
\label{tab:anet_main}
\end{table}

\textbf{ActivityNet-1.3.} As shown in Table~\ref{tab:anet_main}, our method achieves competitive performance on ActivityNet-1.3. Due to the $200$ action categories in ActivityNet-1.3, existing methods struggle with action classification, making it challenging to achieve optimal detection performance. Consequently, most methods~\cite{actionformer, tridet, adsformer} use classification results from external classifiers like UntrimmedNet~\cite{untrimmednets}. Under this setup, we first report the performance of our method. As shown in Table~\ref{tab:anet_main}, using external classifiers, our method surpasses the corresponding baselines in both I3D~\cite{i3d} and TSP~\cite{tsp} features ($35.6\% \, vs. \, 36.8\%$ and $36.6\% \, vs. \, 36.9\%$, respectively).

Thanks to the RCM in our method, which can generate reliable video-level action predictions, our approach reduces the dependency on external classifiers. We denote methods that do not use external classifiers as *\_multi. Additionally, we reproduced ActionFormer\_multi for comparison with our method. As shown in Table~\ref{tab:anet_main}, our method significantly outperforms ActionFormer\_multi with both I3D and TSP features ($30.6\% \, vs. \, 33.7\%$ and $34.2\% \, vs. \, 35.4\%$, respectively). When using InternVideoV2-6B features, our method also outperforms ActionFormer\_multi ($40.4\% \, vs. \, 41.1\%$) and achieves performance comparable to ActionFormer using external classifiers ($41.2\% \, vs. \, 41.1\%$). These experimental results demonstrate that our method can effectively reduce the reliance on external classifiers in existing methods. Furthermore, it is noteworthy that we also implemented TriDet\_multi, which performs significantly worse than ActionFormer\_multi ($38.4\% \, vs. \, 40.4\%$). We speculate that this discrepancy is mainly due to TriDet requiring finer parameter tuning to achieve optimal performance.

\begin{table}[t!]
    \caption{Temporal action detection results on MultiTHUMOS. $^\dagger$ indicates our reproduction. Avg.: average mAP under tIoU thresholds $[0.1:0.1:0.9]$.}
    \centering
    \resizebox{0.8\linewidth}{!}{\begin{tabular}{c|c|cccc}
\hline
\multirow{2}{*}{\textbf{Model}} & \multirow{2}{*}{\textbf{Venue}} & \multicolumn{4}{c}{\textbf{mAP@tIoU}} \\ \cline{3-6} 
 &  & \textbf{0.2} & \textbf{0.5} & \textbf{0.7} & \textbf{Avg.} \\ \hline \hline
TadTR~\cite{tadtr} & TIP'22 & - & 29.1 & - & 27.4 \\
ActionFormer~\cite{actionformer} & ECCV'22 & 46.4 & 32.4 & 15.0 & 28.6 \\
ActionFormer$^\dagger$~\cite{actionformer} & ECCV'22 & 54.0 & 39.8 & 21.2 & 34.8 \\
TemporalMaxer~\cite{temporalmaxer} & arXiv'23 & 47.5 & 33.4 & 17.4 & 29.9 \\
TriDet~\cite{tridet} & CVPR'23 & 49.1 & 34.3 & 17.8 & 30.7 \\
DualDETR~\cite{dualdetr} & CVPR'24 & - & 35.2 & - & 32.6 \\
ADSFormer~\cite{adsformer} & TMM'24 & 55.3 & 40.9 & 21.9 & 35.6 \\
ours & - & 55.4 & 40.7 & 22.1 & 35.7 \\ \hline
\end{tabular}}
\label{tab:multithumos_main}
\end{table}

\textbf{MultiTHUMOS.} To validate the detection performance of our method when multiple action categories appear in a single video, we conducted experiments on the MultiTHUMOS dataset, and the results are shown in Table~\ref{tab:multithumos_main}. From Table~\ref{tab:multithumos_main}, it can be observed that even on challenging datasets like MultiTHUMOS, which contain videos with multiple complex actions (an average of 10.5 action categories per video), our method consistently achieves stable performance improvements. Additionally, from Tables~\ref{tab:multithumos_main} and Table~\ref{tab:complexity}, it can be concluded that although ADSFormer shows performance similar to our method on MultiTHUMOS, the FLOPs and Parameters of ADSFormer are inferior to our method.

\begin{table}[t!]
    \caption{Temporal action detection results of our method on THUMOS'14 using different features and backbones. $^\dagger$ indicates our reproduction. Avg.: average mAP under tIoU thresholds $[0.3:0.1:0.7]$.}
    \centering
    \resizebox{0.95\linewidth}{!}{\begin{tabular}{c|c|cccccc}
\hline
 &  & \multicolumn{6}{c}{\textbf{mAP@tIoU(\%)}} \\ \cline{3-8} 
\multirow{-2}{*}{\textbf{Feature}} & \multirow{-2}{*}{\textbf{Model}} & \textbf{0.3} & \textbf{0.4} & \textbf{0.5} & \textbf{0.6} & \textbf{0.7} & \textbf{Avg.} \\ \hline \hline
 & ActionFormer$_{ECCV'22}$ & 82.1 & 77.8 & 71.0 & 59.4 & 43.9 & 66.8 \\
 & \cellcolor[HTML]{C0C0C0}ActionFormer+ours & \cellcolor[HTML]{C0C0C0}84.0 & \cellcolor[HTML]{C0C0C0}80.3 & \cellcolor[HTML]{C0C0C0}72.7 & \cellcolor[HTML]{C0C0C0}61.8 & \cellcolor[HTML]{C0C0C0}46.9 & \cellcolor[HTML]{C0C0C0}69.1 \\
 & TriDet$_{CVPR'23}$ & 83.6 & 80.1 & 72.9 & 62.4 & 47.4 & 69.3 \\
\multirow{-4}{*}{I3D} & \cellcolor[HTML]{C0C0C0}TriDet+ours & \cellcolor[HTML]{C0C0C0}83.8 & \cellcolor[HTML]{C0C0C0}80.2 & \cellcolor[HTML]{C0C0C0}72.9 & \cellcolor[HTML]{C0C0C0}62.5 & \cellcolor[HTML]{C0C0C0}48.1 & \cellcolor[HTML]{C0C0C0}69.5 \\ \hline
 & ActionFormer$_{ECCV'22}$ & 84.0 & 79.6 & 73.0 & 63.5 & 47.7 & 69.6 \\
 & \cellcolor[HTML]{C0C0C0}ActionFormer+ours & \cellcolor[HTML]{C0C0C0}85.8 & \cellcolor[HTML]{C0C0C0}81.3 & \cellcolor[HTML]{C0C0C0}74.6 & \cellcolor[HTML]{C0C0C0}63.9 & \cellcolor[HTML]{C0C0C0}50.8 & \cellcolor[HTML]{C0C0C0}71.3 \\
 & TriDet$_{CVPR'23}$ & 84.8 & 80.0 & 73.3 & 63.8 & 48.8 & 70.1 \\
\multirow{-4}{*}{VideoMAE V2} & \cellcolor[HTML]{C0C0C0}TriDet+ours & \cellcolor[HTML]{C0C0C0}86.8 & \cellcolor[HTML]{C0C0C0}82.4 & \cellcolor[HTML]{C0C0C0}75.3 & \cellcolor[HTML]{C0C0C0}64.4 & \cellcolor[HTML]{C0C0C0}50.6 & \cellcolor[HTML]{C0C0C0}71.9 \\ \hline
 & ActionFormer$_{ECCV'22}$ & 82.3 & 81.9 & 75.1 & 65.8 & 50.3 & 71.1 \\
 & \cellcolor[HTML]{C0C0C0}ActionFormer+ours & \cellcolor[HTML]{C0C0C0}86.8 & \cellcolor[HTML]{C0C0C0}82.6 & \cellcolor[HTML]{C0C0C0}76.1 & \cellcolor[HTML]{C0C0C0}65.3 & \cellcolor[HTML]{C0C0C0}51.3 & \cellcolor[HTML]{C0C0C0}72.4 \\
 & TriDet$_{CVPR'23}$ & 87.5 & 83.8 & 77.4 & 67.1 & 51.9 & 73.5 \\
\multirow{-4}{*}{InternVideoV2-6B} & \cellcolor[HTML]{C0C0C0}TriDet+ours & \cellcolor[HTML]{C0C0C0}88.6 & \cellcolor[HTML]{C0C0C0}83.8 & \cellcolor[HTML]{C0C0C0}78.0 & \cellcolor[HTML]{C0C0C0}66.8 & \cellcolor[HTML]{C0C0C0}51.2 & \cellcolor[HTML]{C0C0C0}73.7 \\ \hline
 & AdaTAD$_{CVPR'24}$ & 86.0 & 81.9 & 75.0 & 63.3 & 49.7 & 71.1 \\
 & AdaTAD$^\dagger$ & 85.3 & 81.2 & 74.6 & 63.3 & 49.4 & 70.8 \\
\multirow{-3}{*}{VideoMAE-B} & \cellcolor[HTML]{C0C0C0}AdaTAD+ours & \cellcolor[HTML]{C0C0C0}86.2 & \cellcolor[HTML]{C0C0C0}82.1 & \cellcolor[HTML]{C0C0C0}74.9 & \cellcolor[HTML]{C0C0C0}63.5 & \cellcolor[HTML]{C0C0C0}49.7 & \cellcolor[HTML]{C0C0C0}71.3 \\ \hline
\end{tabular}}
\label{tab:thumos_diff_feature_cmp}
\end{table}

\textbf{Robustness.} To validate the robustness of our method, we conducted experiments on various features such as I3D, VideoMAE-B, VideoMAE V2, and InternVideoV2-6B, as well as different backbones including ActionFormer, TriDet, and AdaTAD. We provide all relevant experiment configurations, training logs, and pre-trained models on \href{https://github.com/zhenyingfang/BDRC-Net/tree/opentad}{Github}. As shown in Table~\ref{tab:thumos_diff_feature_cmp}, our method consistently improves performance across different backbones when using the same features. Meanwhile, as shown in Table~\ref{tab:complexity}, our method demonstrates good robustness while having a comparative advantage in terms of FLOPs and Parameters compared to TriDet and ADSFormer.

\subsection{Ablation Studies}\label{subsec:ablation_studies}

In this section, we conduct multiple ablation studies on the THUMOS'14 dataset. These studies include comparisons with other mixed methods, an analysis of the effectiveness of each module in BDRC-Net, and an investigation of the impact of various hyper-parameters. Notably, we first implement a BDRC-Net baseline based on A2Net's MSB, the state-of-the-art mixed method. We then compare the performance of BDRC-Net with A2Net to verify the advantage of BDRC-Net over other mixed methods. Subsequently, we conduct further ablation studies based on this baseline.


\textbf{Comparison with other mixed methods.} To fairly validate the effectiveness of our method, we implement a BDRC-Net baseline based on A2Net's MSB. Moreover, we compare its performance with that of A2Net, which is currently the best-performing mixed method. As shown in Table~\ref{tab:main_components}, our reproduce A2Net exhibited better performance than the official implementation ($Row \, 1 \: vs. \: Row \, 2$). Moreover, our method significantly outperforms A2Net when only using BDM in BDRC-Net, with an improvement of $3.2\%$ in average mAP ($Row \, 2 \: vs. \: Row \, 6$). When using both BDM and RCM, the performance improves due to the reliable classification predictions provided by RCM, which can reduce false positives in action category predictions. The average mAP improve by $4.2\%$ compared to A2Net ($Row \, 2 \: vs. \: Row \, 7$).

\begin{table}[t!]
    \caption{Ablation studies on the effectiveness of each part of our framework on THUMOS'14, including CCSM, RRSM, and RCM. A2Net$^\dagger$: Our reproduce.}
    \centering
    \resizebox{1.0\linewidth}{!}{\begin{tabular}{c|ccc|ccc|c}
\hline
\multirow{2}{*}{Method} & \multirow{2}{*}{CCSM} & \multirow{2}{*}{RRSM} & \multirow{2}{*}{RCM} & \multicolumn{3}{c|}{mAP@tIoU(\%)} & \multirow{2}{*}{\begin{tabular}[c]{@{}c@{}}average mAP(\%)\\ $[0.5:0.1:0.9]$\end{tabular}} \\
 &  &  &  & 0.5 & 0.7 & 0.9 &  \\ \hline \hline
A2Net &  &  &  & 45.5 & 17.2 & - & - \\
A2Net$^\dagger$ &  &  &  & 45.2 & 19.1 & 0.9 & 20.9 \\ \hline
\multirow{5}{*}{BDRC-Net} &  &  & $\surd$ & 46.4 & 19.4 & 0.9 & 21.4 \\
 & $\surd$ &  &  & 46.7 & 22.1 & 1.2 & 23.0 \\
 & $\surd$ &  & $\surd$ & 50.0 & 23.4 & 1.3 & 24.4 \\
 & $\surd$ & $\surd$ &  & 47.2 & 23.8 & 1.9 & 24.1 \\
 & $\surd$ & $\surd$ & $\surd$ & 50.6 & 25.2 & 2.0 & 25.7 \\ \hline
\end{tabular}}
\label{tab:main_components}
\end{table}

\textbf{Main components analysis.} We demonstrate the effectiveness of our proposed components in BDRC-Net: BDM and RCM. Moreover, our experiments are based on A2Net's MSB. Specifically, BDM consists of CCSM and RRSM, and we have individually validated the efficacy of these two sub-modules. As described in Table~\ref{tab:main_components}, CCSM yields a $1.1\%$ absolute improvement in average mAP ($Row \, 4$). This improvement primarily stems from CCSM's ability to locate coarse boundaries of actions through boundary discretization stably. Consequently, CCSM exhibits minimal performance gain at higher tIoU thresholds. To refine the coarse results obtained by CCSM, RRSM is introduced. RRSM offers a $1.1\%$ absolute improvement in average mAP compared to CCSM ($Row \, 4 \, vs. \, Row \, 6$). Especially, RRSM significantly enhances at higher tIoU thresholds, such as a $1.7\%$ and $0.7\%$ improvement in $mAP@tIoU=0.7$ and $mAP@tIoU=0.9$, respectively. This demonstrates RRSM's capability to predict more precise action boundaries. Additionally, our RCM is employed to predict reliable video-level action categories for filtering false positives in snippet-level action predictions. Therefore, RCM can easily integrate into other action detectors. When integrated into A2Net, RCM contributes a $0.5\%$ absolute improvement in average mAP ($Row \, 2 \, vs. \, Row \, 3$). Additionally, RCM yields average mAP improvements of $1.4\%$ for CCSM and $1.6\%$ for BDM, respectively ($Row \, 5$ and $Row \, 7$). The consistent and stable enhancements across different models robustly attest to the effectiveness of RCM.

\textbf{Hyper-parameters.} As mentioned in Sec.~\ref{subsec:implementation_details}, all the hyper-parameters in BDRC-Net are set through empirical search. In this section, we delve into the discussion on determining these hyper-parameters and their respective impact. All experiments are conducted on the THUMOS '14 dataset using A2Net's MSB.

\begin{table}[t!]
    \caption{Ablation studies of the hyper-parameters on THUMOS'14 include $\lambda_{rs}$, $\lambda_{norm}$, and $\sigma$, which are used for label assignment in RRSM, scaling loss, and label assignment in CCSM, respectively.}
    \centering
    \resizebox{0.85\linewidth}{!}{\begin{tabular}{c|c|ccc|c}
\hline
\multirow{2}{*}{Parameter} & \multirow{2}{*}{Value} & \multicolumn{3}{c|}{mAP@tIoU(\%)} & \multirow{2}{*}{\begin{tabular}[c]{@{}c@{}}average mAP(\%)\\ $[0.5:0.1:0.9]$\end{tabular}} \\
 &  & 0.5 & 0.7 & 0.9 &  \\ \hline \hline
\multirow{4}{*}{$\lambda_{rs}$} & 0.4 & 49.6 & 25.3 & 1.9 & 25.4 \\
 & 0.5 & 50.6 & 25.2 & 2.0 & 25.7 \\
 & 0.6 & 48.0 & 25.0 & 2.0 & 24.9 \\
 & 0.7 & 46.1 & 21.8 & 1.3 & 22.9 \\ \hline \hline
\multirow{3}{*}{$\lambda_{norm}$} & 80 & 47.7 & 24.8 & 1.8 & 24.4 \\
 & 90 & 50.6 & 25.2 & 2.0 & 25.7 \\
 & 100 & 50.4 & 24.9 & 1.8 & 25.5 \\ \hline \hline
\multirow{3}{*}{$\sigma$} & $\sqrt{0.1}$ & 48.5 & 23.9 & 1.8 & 24.5 \\
 & $\sqrt{0.2}$ & 50.6 & 25.2 & 2.0 & 25.7 \\
 & $\sqrt{0.3}$ & 49.1 & 25.4 & 1.8 & 25.0 \\ \hline
\end{tabular}}
\label{tab:parameters}
\end{table}

As mentioned in Sec.~\ref{subsec:training}, the parameter $\lambda_{rs}$ is used to assign labels for RRSM. As the experimental results presented in Table~\ref{tab:parameters} indicates, the best detection performance is achieved when taking an intermediate value for $\lambda_{rs}$. When $\lambda_{rs}$ is decreased, it introduces many hard examples during RRSM loss computation, increasing the learning difficulty and declining performance. On the other hand, increasing $\lambda_{rs}$ reduces the number of positive examples in RRSM, also leading to decreased performance.

In BDRC-Net, $\sigma$ is used to assign labels for CCSM. At the same time, when calculating the loss function, we scale the loss using loss normalization $\lambda_{norm}$ followed by ActionFormer~\cite{actionformer}. As shown in the experimental results in Table~\ref{tab:parameters}, BDRC-Net achieves optimal performance when $\sigma$ and $\lambda_{norm}$ are set to $\sqrt{0.2}$ and $90$, respectively.

\begin{table}[t!]
    \caption{Ablation studies of the number of bins and bin coverage on THUMOS'14.}
    \centering
    \resizebox{0.8\linewidth}{!}{\begin{tabular}{c|c|ccc|c}
\hline
\multirow{2}{*}{$W$} & \multirow{2}{*}{$b$} & \multicolumn{3}{c|}{mAP@tIoU(\%)} & \multirow{2}{*}{\begin{tabular}[c]{@{}c@{}}average mAP(\%)\\ $[0.5:0.1:0.9]$\end{tabular}} \\
 &  & 0.5 & 0.7 & 0.9 &  \\ \hline \hline
\multirow{2}{*}{4} & 0.25 & 48.1 & 24.4 & 1.8 & 24.4 \\
 & 0.5 & 49.9 & 25.0 & 1.8 & 25.5 \\ \hline \hline
\multirow{2}{*}{8} & 0.25 & 50.6 & 25.2 & 2.0 & 25.7 \\
 & 0.5 & 48.1 & 22.9 & 1.6 & 23.9 \\ \hline
\end{tabular}}
\label{tab:bin_set}
\end{table}

\textbf{Ablation on the number and coverage length of bins.} We present the ablation results of the number and coverage range of bins in BDM, conducted on the THUMOS'14 dataset using A2Net's MSB. Table~\ref{tab:bin_set} shows that the best result is achieved with $W=8$ and $b=0.25$. We also find that when $W=4$, a value of $b=0.5$ yielded better performance than $b=0.25$. This is because, in A2Net, the maximum length of the ground truth in the last MSB's layer is $2$. When $W \times b$ equals this length, all bins can encompass all possible boundaries, improving performance. When $W \times b$ exceeds this length, although it still covers all possible boundaries, the increased coverage length per bin decreases the accuracy of CCSM boundary predictions. Similarly, when $W \times b$ is less than the maximum ground truth length, bins cannot cover all action boundaries. While reducing the coverage length per bin enhances the accuracy of predicting shorter ground truth boundaries, it leads to poorer boundary prediction performance when the ground truth length exceeds $W \times b$, resulting in a performance decrease.

\begin{table}[t!]
    \caption{Ablation studies of confidence score on THUMOS'14. $score \, 1$: $P_{sc}(t_i,k_i)$; $score \, 2$: $P_{sc}(t_i,k_i) \times \max P_{cs}(t_i)$; $score \, 3$: $P_{sc}(t_i,k_i) \times \max P_{ce}(t_i)$; $score \, 4$: $P_{sc}(t_i,k_i) \times \max P_{cs}(t_i) \times \max P_{ce}(t_i)$; $score \, 5$: $P_{sc}(t_i,k_i) \times \sqrt{\max P_{cs}(t_i) \times \max P_{ce}(t_i)}$.}
    \centering
    \resizebox{0.75\linewidth}{!}{\begin{tabular}{c|ccc|c}
\hline
\multirow{2}{*}{Method} & \multicolumn{3}{c|}{mAP@tIoU(\%)} & \multirow{2}{*}{\begin{tabular}[c]{@{}c@{}}average mAP(\%)\\ $[0.5:0.1:0.9]$\end{tabular}} \\
 & 0.5 & 0.7 & 0.9 &  \\ \hline \hline
$score \, 1$ & 50.2 & 24.4 & 1.9 & 25.2 \\
$score \, 2$ & 50.2 & 24.4 & 2.0 & 25.4 \\
$score \, 3$ & 50.5 & 24.7 & 1.9 & 25.5 \\
$score \, 4$ & 50.5 & 24.9 & 2.0 & 25.6 \\
$score \, 5$ & 50.6 & 25.2 & 2.0 & 25.7 \\ \hline
\end{tabular}}
\label{tab:score_fusion}
\end{table}

\textbf{Confidence score.} In Table~\ref{tab:score_fusion}, we present the impact of different confidence score calculation methods on detection performance. Accurate confidence scoring leads to a more precise ranking of detection results and is beneficial for practical applications of action detection. We observe that considering both snippet-level classification scores and the predicted probabilities of bins in BDM leads to better confidence scores ($score \, 4$ and $score \, 5$). Furthermore, achieving optimal performance involves reducing the weight of bin prediction probabilities in the confidence score calculation ($score \, 5$). In contrast, $score \, 1$, $score \, 2$, and $score \, 3$ only consider partial prediction probabilities, which reduces the quality of confidence scores and decreases performance.

\subsection{Analysis}\label{subsec:analysis}

In this section, we analyze the model efficiency of BDRC-Net, the classification performance of RCM, and finally, provide visible results to demonstrate the precision of boundary predictions for BDRC-Net.

\begin{table}[t!]
    \caption{Comparison of FLOPs and Parameters on THUMOS'14.}
    \centering
    \resizebox{0.8\linewidth}{!}{\begin{tabular}{c|cc|c}
\hline
\multirow{2}{*}{\textbf{Method}} & \multicolumn{2}{c|}{\textbf{Detection Head}} & \textbf{mAP(\%)} \\ \cline{2-4} 
 & \textbf{FLOPs(G)} & \textbf{Params(M)} & \textbf{Avg.} \\ \hline \hline
ActionFormer & 14.5 & 3.2 & 66.8 \\
TemporalMaxer & 14.5 & 3.2 & 67.7 \\
TriDet & 29.3 & 6.4 & 69.3 \\
ADSFormer & 29.3 & 6.4 & 69.5 \\
ours & 22.6 & 5.0 & 69.5 \\ \hline
\end{tabular}}
\label{tab:complexity}
\end{table}

\textbf{Model Efﬁciency.} We compared the FLOPs and Params of our method with those of existing classical methods, as shown in Table~\ref{tab:complexity}. Table~\ref{tab:complexity} shows that although TriDet and ADSFormer achieved performance close to ours, their FLOPs and Params are higher than our method. Additionally, while our method has higher FLOPs and Params than ActionFormer and TemporalMaxer, our method's performance is significantly better than these methods. Most importantly, as shown in Table~\ref{tab:anet_main}, on datasets with a large number of action categories, such as ActivityNet-1.3, our method can significantly reduce the reliance on external classifiers in existing methods, potentially leading to a substantial reduction in inference time.

\begin{table}[t!]
    \caption{Compare the $F1$ score of action categories in the detection results of different models on the THUMOS'14 dataset. A2Net$^\dagger$: Our reproduce.}
    \centering
    \resizebox{0.9\linewidth}{!}{\begin{tabular}{c|ccccc}
\hline
Method & A2Net & A2Net$^\dagger$ & \begin{tabular}[c]{@{}c@{}}A2Net$^\dagger$\\ w/ RCM\end{tabular} & \begin{tabular}[c]{@{}c@{}}BDRC-Net\\ w/o RCM\end{tabular} & BDRC-Net \\ \hline \hline
F1 & 0.18 & 0.17 & 0.71 & 0.25 & 0.79 \\ \hline
\end{tabular}}
\label{tab:rcm_analysis}
\end{table}

\begin{figure*}[!t]
\centering
\subfloat[]{\includegraphics[width=5.3in]{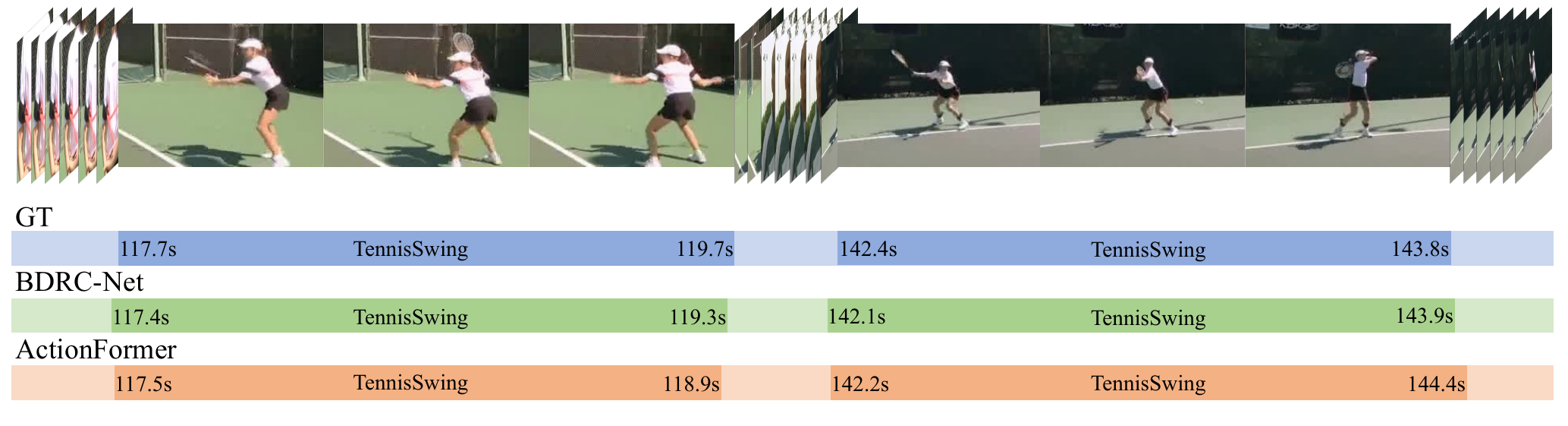}%
\label{subfig:TennisSwing}}
\hfil
\subfloat[]{\includegraphics[width=5.3in]{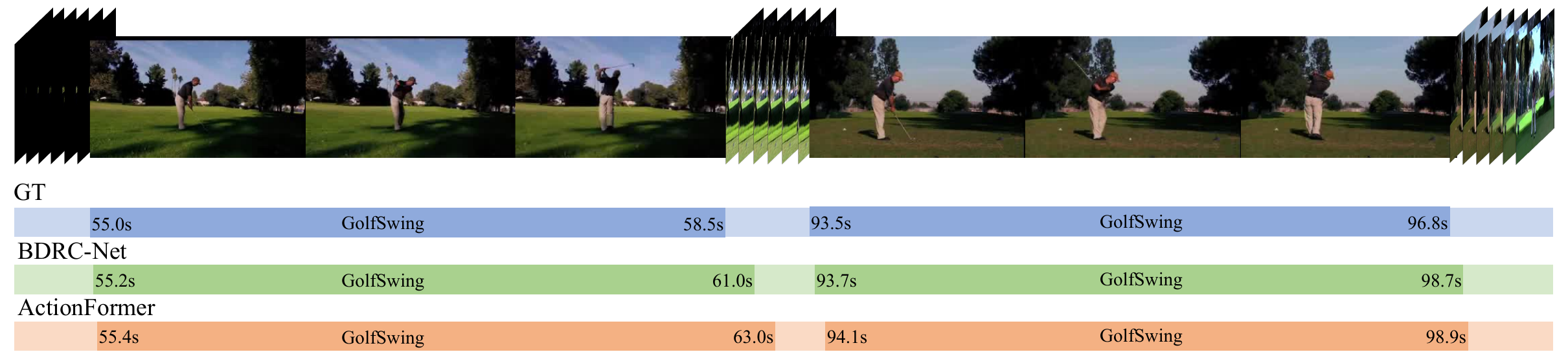}%
\label{subfig:GolfSwing}}
\hfill
\subfloat[]{\includegraphics[width=5.3in]{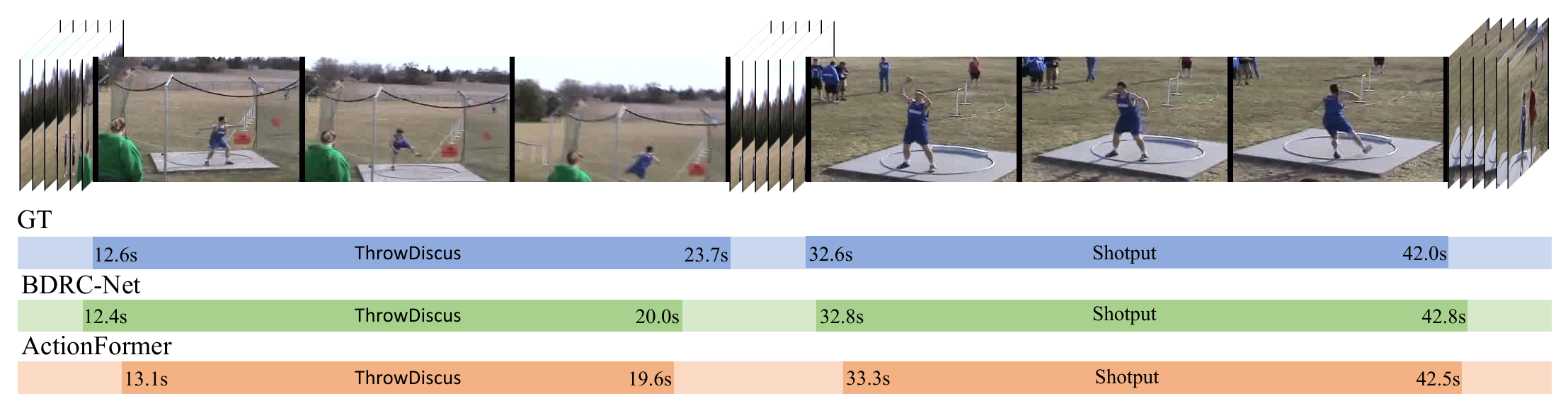}%
\label{subfig:ThrowDiscus_and_Shotput}}
\hfil
\subfloat[]{\includegraphics[width=5.3in]{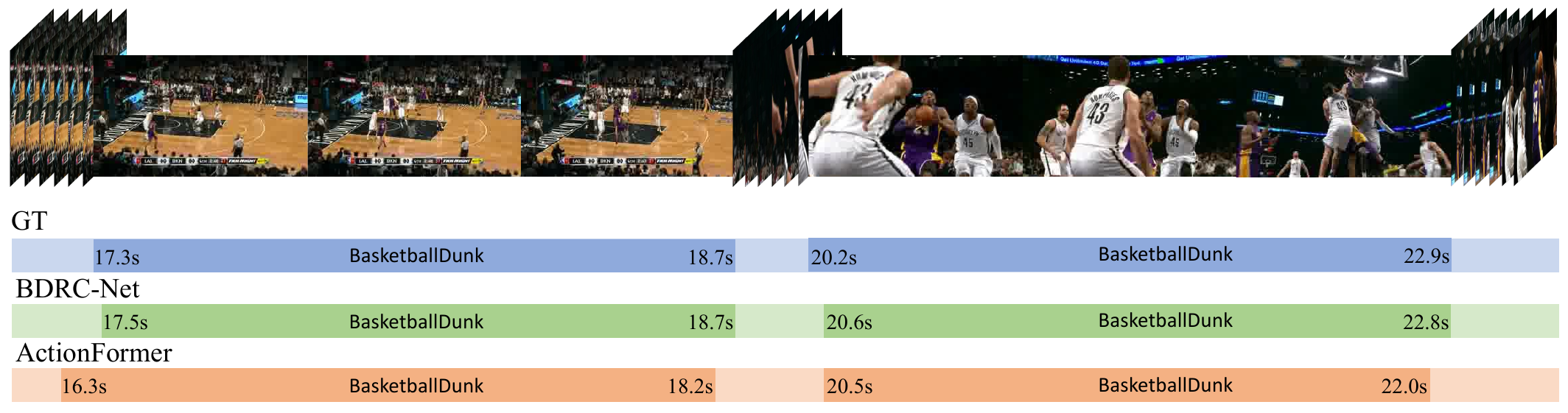}%
\label{subfig:BasketballDunk}}
\caption{Visualizations of the action detection results for ActionFormer and BDRC-Net on THUMOS'14 dataset. (a) TennisSwing. (b) GolfSwing. (c) ThrowDiscus and Shotput. (d) BasketballDunk.}
\label{fig:viz}
\end{figure*}

\textbf{RCM analysis.} To verify whether RCM produces reliable classification results, we compare the $F1$ scores of action classification among A2Net, A2Net$^\dagger$, A2Net$^\dagger$ with RCM, BDRC-Net without RCM, and BDRC-Net. The experimental results are shown in Table~\ref{tab:rcm_analysis}. It can be observed that when without RCM, both A2Net and BDRC-Net have lower $F1$ scores, demonstrating that the classification performance is poor without RCM. When RCM is added, it provides reliable video-level action predictions and filters out a large number of false positives in their prediction results. As a result, the $F1$ scores of both A2Net and BDRC-Net significantly increase.

\textbf{Visualization results.} To demonstrate the improvement of our framework, we present some examples in Fig.~\ref{fig:viz} that compare our method with ActionFormer, on the THUMOS'14 dataset. It can be observed that our method can generate highly precise and reliable action boundaries, and the optimized temporal boundaries correspond better to actual action instances. Furthermore, our method performs well in detecting action instances within scenarios involving multiple action categories (Fig.~\ref{subfig:ThrowDiscus_and_Shotput}) and adapts effectively to various complex contexts (Fig.~\ref{subfig:BasketballDunk}).

\section{Conclusion}
In this paper, we propose a novel method, BDRC-Net, for temporal action detection task. BDRC-Net consists of a boundary discretization module (BDM) and a reliable classification module (RCM). BDM elegantly integrates anchor-based and anchor-free approaches through boundary discretization, achieving a superior mixed method that mitigates brute-force merging and handcrafted anchors design in existing mixed methods. In detail, BDM first predicts stable but coarse action boundaries using a coarse classification sub-module (CCSM). Then, the refined regression sub-module (RRSM) is used to refine the coarse results obtained from CCSM, generating more precise action boundaries. Moreover, RCM filters false positives in snippet-level action category predictions by predicting reliable video-level action categories, thereby enhancing the accuracy of the detection results. Extensive experiments conducted on different benchmarks demonstrate that our proposed method achieves competitive detection performance.

\bibliographystyle{IEEEtran}
\bibliography{BDRC-Net}

\begin{thebibliography}{10}
\providecommand{\url}[1]{#1}
\csname url@samestyle\endcsname
\providecommand{\newblock}{\relax}
\providecommand{\bibinfo}[2]{#2}
\providecommand{\BIBentrySTDinterwordspacing}{\spaceskip=0pt\relax}
\providecommand{\BIBentryALTinterwordstretchfactor}{4}
\providecommand{\BIBentryALTinterwordspacing}{\spaceskip=\fontdimen2\font plus
\BIBentryALTinterwordstretchfactor\fontdimen3\font minus \fontdimen4\font\relax}
\providecommand{\BIBforeignlanguage}[2]{{%
\expandafter\ifx\csname l@#1\endcsname\relax
\typeout{** WARNING: IEEEtran.bst: No hyphenation pattern has been}%
\typeout{** loaded for the language `#1'. Using the pattern for}%
\typeout{** the default language instead.}%
\else
\language=\csname l@#1\endcsname
\fi
#2}}
\providecommand{\BIBdecl}{\relax}
\BIBdecl

\bibitem{videosummary}
Q.~Li, J.~Chen, Q.~Xie, and X.~Han, ``Video summarization for event-centric videos,'' \emph{Neural Networks}, 2023.

\bibitem{surveillance1}
H.~Zhang and C.-W. Ngo, ``A fine granularity object-level representation for event detection and recounting,'' \emph{IEEE Transactions on Multimedia}, vol.~21, no.~6, pp. 1450--1463, 2018.

\bibitem{behavior1}
K.~Kumar and D.~D. Shrimankar, ``F-des: Fast and deep event summarization,'' \emph{IEEE Transactions on Multimedia}, vol.~20, no.~2, pp. 323--334, 2017.

\bibitem{behavior2}
S.~Ma, J.~Zhang, S.~Sclaroff, N.~Ikizler-Cinbis, and L.~Sigal, ``Space-time tree ensemble for action recognition and localization,'' \emph{International Journal of Computer Vision}, vol. 126, pp. 314--332, 2018.

\bibitem{ssad}
T.~Lin, X.~Zhao, and Z.~Shou, ``Single shot temporal action detection,'' in \emph{Proceedings of the 25th ACM international conference on Multimedia}, 2017, pp. 988--996.

\bibitem{rc3d}
H.~Xu, A.~Das, and K.~Saenko, ``R-c3d: Region convolutional 3d network for temporal activity detection,'' in \emph{Proceedings of the IEEE international conference on computer vision}, 2017, pp. 5783--5792.

\bibitem{pbrnet}
Q.~Liu and Z.~Wang, ``Progressive boundary refinement network for temporal action detection,'' in \emph{Proceedings of the AAAI conference on artificial intelligence}, vol.~34, no.~07, 2020, pp. 11\,612--11\,619.

\bibitem{tbos}
Z.~Li and L.~Yao, ``Three birds with one stone: Multi-task temporal action detection via recycling temporal annotations,'' in \emph{Proceedings of the IEEE/CVF Conference on Computer Vision and Pattern Recognition}, 2021, pp. 4751--4760.

\bibitem{tallformer}
F.~Cheng and G.~Bertasius, ``Tallformer: Temporal action localization with a long-memory transformer,'' in \emph{Computer Vision--ECCV 2022: 17th European Conference, Tel Aviv, Israel, October 23--27, 2022, Proceedings, Part XXXIV}.\hskip 1em plus 0.5em minus 0.4em\relax Springer, 2022, pp. 503--521.

\bibitem{ssn}
Y.~Zhao, Y.~Xiong, L.~Wang, Z.~Wu, X.~Tang, and D.~Lin, ``Temporal action detection with structured segment networks,'' in \emph{Proceedings of the IEEE international conference on computer vision}, 2017, pp. 2914--2923.

\bibitem{bsn}
T.~Lin, X.~Zhao, H.~Su, C.~Wang, and M.~Yang, ``Bsn: Boundary sensitive network for temporal action proposal generation,'' in \emph{Proceedings of the European conference on computer vision (ECCV)}, 2018, pp. 3--19.

\bibitem{afsd}
C.~Lin, C.~Xu, D.~Luo, Y.~Wang, Y.~Tai, C.~Wang, J.~Li, F.~Huang, and Y.~Fu, ``Learning salient boundary feature for anchor-free temporal action localization,'' in \emph{Proceedings of the IEEE/CVF Conference on Computer Vision and Pattern Recognition}, 2021, pp. 3320--3329.

\bibitem{brem}
J.~Hu, L.~Zhuang, B.~Wang, T.~Ge, Y.~Jiang, H.~Li \emph{et~al.}, ``Estimation of reliable proposal quality for temporal action detection,'' \emph{arXiv preprint arXiv:2204.11695}, 2022.

\bibitem{actionformer}
C.-L. Zhang, J.~Wu, and Y.~Li, ``Actionformer: Localizing moments of actions with transformers,'' in \emph{European Conference on Computer Vision}.\hskip 1em plus 0.5em minus 0.4em\relax Springer, 2022, pp. 492--510.

\bibitem{rtdnet}
J.~Tan, J.~Tang, L.~Wang, and G.~Wu, ``Relaxed transformer decoders for direct action proposal generation,'' in \emph{Proceedings of the IEEE/CVF international conference on computer vision}, 2021, pp. 13\,526--13\,535.

\bibitem{tadtr}
X.~Liu, Q.~Wang, Y.~Hu, X.~Tang, S.~Zhang, S.~Bai, and X.~Bai, ``End-to-end temporal action detection with transformer,'' \emph{IEEE Transactions on Image Processing}, vol.~31, pp. 5427--5441, 2022.

\bibitem{react}
D.~Shi, Y.~Zhong, Q.~Cao, J.~Zhang, L.~Ma, J.~Li, and D.~Tao, ``React: Temporal action detection with relational queries,'' in \emph{European conference on computer vision}.\hskip 1em plus 0.5em minus 0.4em\relax Springer, 2022, pp. 105--121.

\bibitem{selfdetr}
J.~Kim, M.~Lee, and J.-P. Heo, ``Self-feedback detr for temporal action detection,'' in \emph{Proceedings of the IEEE/CVF International Conference on Computer Vision}, 2023, pp. 10\,286--10\,296.

\bibitem{dualdetr}
Y.~Zhu, G.~Zhang, J.~Tan, G.~Wu, and L.~Wang, ``Dual detrs for multi-label temporal action detection,'' \emph{arXiv preprint arXiv:2404.00653}, 2024.

\bibitem{tetad}
H.-J. Kim, J.-H. Hong, H.~Kong, and S.-W. Lee, ``Te-tad: Towards full end-to-end temporal action detection via time-aligned coordinate expression,'' \emph{arXiv preprint arXiv:2404.02405}, 2024.

\bibitem{pcad}
S.~Zhu, X.~Yang, J.~Yu, Z.~Fang, M.~Wang, and Q.~Huang, ``Proposal complementary action detection,'' \emph{ACM Transactions on Multimedia Computing, Communications, and Applications (TOMM)}, vol.~16, no.~2s, pp. 1--12, 2020.

\bibitem{mgg}
Y.~Liu, L.~Ma, Y.~Zhang, W.~Liu, and S.-F. Chang, ``Multi-granularity generator for temporal action proposal,'' in \emph{Proceedings of the IEEE/CVF conference on computer vision and pattern recognition}, 2019, pp. 3604--3613.

\bibitem{a2net}
L.~Yang, H.~Peng, D.~Zhang, J.~Fu, and J.~Han, ``Revisiting anchor mechanisms for temporal action localization,'' \emph{IEEE Transactions on Image Processing}, vol.~29, pp. 8535--8548, 2020.

\bibitem{untrimmednets}
L.~Wang, Y.~Xiong, D.~Lin, and L.~Van~Gool, ``Untrimmednets for weakly supervised action recognition and detection,'' in \emph{Proceedings of the IEEE conference on Computer Vision and Pattern Recognition}, 2017, pp. 4325--4334.

\bibitem{he2022asm}
B.~He, X.~Yang, L.~Kang, Z.~Cheng, X.~Zhou, and A.~Shrivastava, ``Asm-loc: Action-aware segment modeling for weakly-supervised temporal action localization,'' in \emph{Proceedings of the IEEE/CVF conference on computer vision and pattern recognition}, 2022, pp. 13\,925--13\,935.

\bibitem{zhao2023novel}
Y.~Zhao, H.~Zhang, Z.~Gao, W.~Gao, M.~Wang, and S.~Chen, ``A novel action saliency and context-aware network for weakly-supervised temporal action localization,'' \emph{IEEE Transactions on Multimedia}, 2023.

\bibitem{lprfzy}
Z.~Fang, J.~Fan, and J.~Yu, ``Lpr: learning point-level temporal action localization through re-training,'' \emph{Multimedia Systems}, vol.~29, no.~5, pp. 2545--2562, 2023.

\bibitem{decoupssad}
Y.~Huang, Q.~Dai, and Y.~Lu, ``Decoupling localization and classification in single shot temporal action detection,'' in \emph{2019 IEEE International Conference on Multimedia and Expo (ICME)}.\hskip 1em plus 0.5em minus 0.4em\relax IEEE, 2019, pp. 1288--1293.

\bibitem{talnet}
Y.-W. Chao, S.~Vijayanarasimhan, B.~Seybold, D.~A. Ross, J.~Deng, and R.~Sukthankar, ``Rethinking the faster r-cnn architecture for temporal action localization,'' in \emph{proceedings of the IEEE conference on computer vision and pattern recognition}, 2018, pp. 1130--1139.

\bibitem{gtad}
M.~Xu, C.~Zhao, D.~S. Rojas, A.~Thabet, and B.~Ghanem, ``G-tad: Sub-graph localization for temporal action detection,'' in \emph{Proceedings of the IEEE/CVF Conference on Computer Vision and Pattern Recognition}, 2020, pp. 10\,156--10\,165.

\bibitem{fasterrcnn}
S.~Ren, K.~He, R.~Girshick, and J.~Sun, ``Faster r-cnn: Towards real-time object detection with region proposal networks,'' \emph{Advances in neural information processing systems}, vol.~28, 2015.

\bibitem{bmn}
T.~Lin, X.~Liu, X.~Li, E.~Ding, and S.~Wen, ``Bmn: Boundary-matching network for temporal action proposal generation,'' in \emph{Proceedings of the IEEE/CVF international conference on computer vision}, 2019, pp. 3889--3898.

\bibitem{liu2021centerness}
Y.~Liu, J.~Chen, X.~Chen, B.~Deng, J.~Huang, and X.-S. Hua, ``Centerness-aware network for temporal action proposal,'' \emph{IEEE Transactions on Circuits and Systems for Video Technology}, vol.~32, no.~1, pp. 5--16, 2021.

\bibitem{lin2019joint}
T.~Lin, X.~Zhao, and H.~Su, ``Joint learning of local and global context for temporal action proposal generation,'' \emph{IEEE Transactions on Circuits and Systems for Video Technology}, vol.~30, no.~12, pp. 4899--4912, 2019.

\bibitem{eun2019srg}
H.~Eun, S.~Lee, J.~Moon, J.~Park, C.~Jung, and C.~Kim, ``Srg: Snippet relatedness-based temporal action proposal generator,'' \emph{IEEE Transactions on Circuits and Systems for Video Technology}, vol.~30, no.~11, pp. 4232--4244, 2019.

\bibitem{xu2019cascaded}
L.~Xu, X.~Wang, W.~Liu, and B.~Feng, ``Cascaded boundary network for high-quality temporal action proposal generation,'' \emph{IEEE Transactions on Circuits and Systems for Video Technology}, vol.~30, no.~10, pp. 3702--3713, 2019.

\bibitem{ram}
P.~Chen, C.~Gan, G.~Shen, W.~Huang, R.~Zeng, and M.~Tan, ``Relation attention for temporal action localization,'' \emph{IEEE Transactions on Multimedia}, vol.~22, no.~10, pp. 2723--2733, 2019.

\bibitem{nms}
A.~Rosenfeld and M.~Thurston, ``Edge and curve detection for visual scene analysis,'' \emph{IEEE Transactions on computers}, vol. 100, no.~5, pp. 562--569, 1971.

\bibitem{focalloss}
T.-Y. Lin, P.~Goyal, R.~Girshick, K.~He, and P.~Doll{\'a}r, ``Focal loss for dense object detection,'' in \emph{Proceedings of the IEEE international conference on computer vision}, 2017, pp. 2980--2988.

\bibitem{thumos14}
H.~Idrees, A.~R. Zamir, Y.-G. Jiang, A.~Gorban, I.~Laptev, R.~Sukthankar, and M.~Shah, ``The thumos challenge on action recognition for videos “in the wild”,'' \emph{Computer Vision and Image Understanding}, vol. 155, pp. 1--23, 2017.

\bibitem{activitynet1.3}
F.~Caba~Heilbron, V.~Escorcia, B.~Ghanem, and J.~Carlos~Niebles, ``Activitynet: A large-scale video benchmark for human activity understanding,'' in \emph{Proceedings of the IEEE Conference on Computer Vision and Pattern Recognition (CVPR)}, June 2015.

\bibitem{c3d}
D.~Tran, L.~Bourdev, R.~Fergus, L.~Torresani, and M.~Paluri, ``Learning spatiotemporal features with 3d convolutional networks,'' in \emph{Proceedings of the IEEE international conference on computer vision}, 2015, pp. 4489--4497.

\bibitem{p3d}
Z.~Qiu, T.~Yao, and T.~Mei, ``Learning spatio-temporal representation with pseudo-3d residual networks,'' in \emph{proceedings of the IEEE International Conference on Computer Vision}, 2017, pp. 5533--5541.

\bibitem{i3d}
J.~Carreira and A.~Zisserman, ``Quo vadis, action recognition? a new model and the kinetics dataset,'' in \emph{proceedings of the IEEE Conference on Computer Vision and Pattern Recognition}, 2017, pp. 6299--6308.

\bibitem{r21d}
D.~Tran, H.~Wang, L.~Torresani, J.~Ray, Y.~LeCun, and M.~Paluri, ``A closer look at spatiotemporal convolutions for action recognition,'' in \emph{Proceedings of the IEEE conference on Computer Vision and Pattern Recognition}, 2018, pp. 6450--6459.

\bibitem{tsn}
L.~Wang, Y.~Xiong, Z.~Wang, Y.~Qiao, D.~Lin, X.~Tang, and L.~Van~Gool, ``Temporal segment networks: Towards good practices for deep action recognition,'' in \emph{European conference on computer vision}.\hskip 1em plus 0.5em minus 0.4em\relax Springer, 2016, pp. 20--36.

\bibitem{slowfast}
C.~Feichtenhofer, H.~Fan, J.~Malik, and K.~He, ``Slowfast networks for video recognition,'' in \emph{Proceedings of the IEEE/CVF international conference on computer vision}, 2019, pp. 6202--6211.

\bibitem{vgg}
K.~Simonyan and A.~Zisserman, ``Very deep convolutional networks for large-scale image recognition,'' \emph{arXiv preprint arXiv:1409.1556}, 2014.

\bibitem{tong2022videomae}
Z.~Tong, Y.~Song, J.~Wang, and L.~Wang, ``Videomae: Masked autoencoders are data-efficient learners for self-supervised video pre-training,'' \emph{Advances in neural information processing systems}, vol.~35, pp. 10\,078--10\,093, 2022.

\bibitem{wang2023videomae}
L.~Wang, B.~Huang, Z.~Zhao, Z.~Tong, Y.~He, Y.~Wang, Y.~Wang, and Y.~Qiao, ``Videomae v2: Scaling video masked autoencoders with dual masking,'' in \emph{Proceedings of the IEEE/CVF Conference on Computer Vision and Pattern Recognition}, 2023, pp. 14\,549--14\,560.

\bibitem{wang2024internvideo2}
Y.~Wang, K.~Li, X.~Li, J.~Yu, Y.~He, G.~Chen, B.~Pei, R.~Zheng, J.~Xu, Z.~Wang \emph{et~al.}, ``Internvideo2: Scaling video foundation models for multimodal video understanding,'' \emph{arXiv preprint arXiv:2403.15377}, 2024.

\bibitem{gemini}
Y.~Zhou, R.~Wang, H.~Li, and S.-Y. Kung, ``Temporal action localization using long short-term dependency,'' \emph{IEEE Transactions on Multimedia}, vol.~23, pp. 4363--4375, 2020.

\bibitem{stan}
C.~Sun, H.~Song, X.~Wu, Y.~Jia, and J.~Luo, ``Exploiting informative video segments for temporal action localization,'' \emph{IEEE Transactions on Multimedia}, vol.~24, pp. 274--287, 2021.

\bibitem{muses}
X.~Liu, Y.~Hu, S.~Bai, F.~Ding, X.~Bai, and P.~H. Torr, ``Multi-shot temporal event localization: a benchmark,'' in \emph{Proceedings of the IEEE/CVF Conference on Computer Vision and Pattern Recognition}, 2021, pp. 12\,596--12\,606.

\bibitem{vsgn}
C.~Zhao, A.~K. Thabet, and B.~Ghanem, ``Video self-stitching graph network for temporal action localization,'' in \emph{Proceedings of the IEEE/CVF International Conference on Computer Vision}, 2021, pp. 13\,658--13\,667.

\bibitem{yang2022structured}
L.~Yang, J.~Han, T.~Zhao, N.~Liu, and D.~Zhang, ``Structured attention composition for temporal action localization,'' \emph{IEEE Transactions on Image Processing}, 2022.

\bibitem{e2etad}
X.~Liu, S.~Bai, and X.~Bai, ``An empirical study of end-to-end temporal action detection,'' in \emph{Proceedings of the IEEE/CVF Conference on Computer Vision and Pattern Recognition}, 2022, pp. 20\,010--20\,019.

\bibitem{tags}
S.~Nag, X.~Zhu, Y.-Z. Song, and T.~Xiang, ``Proposal-free temporal action detection via global segmentation mask learning,'' in \emph{European Conference on Computer Vision}, 2022, pp. 645--662.

\bibitem{gap}
S.~Nag, X.~Zhu, Y.~Song, and T.~Xiang, ``Post-processing temporal action detection,'' in \emph{Proceedings of the IEEE/CVF Conference on Computer Vision and Pattern Recognition}, 2023, pp. 18\,837--18\,845.

\bibitem{menet}
Z.~Zhao, D.~Wang, and X.~Zhao, ``Movement enhancement toward multi-scale video feature representation for temporal action detection,'' in \emph{Proceedings of the IEEE/CVF International Conference on Computer Vision}, 2023, pp. 13\,555--13\,564.

\bibitem{asliccv23}
J.~Shao, X.~Wang, R.~Quan, J.~Zheng, J.~Yang, and Y.~Yang, ``Action sensitivity learning for temporal action localization,'' \emph{arXiv preprint arXiv:2305.15701}, 2023.

\bibitem{tridet}
D.~Shi, Y.~Zhong, Q.~Cao, L.~Ma, J.~Li, and D.~Tao, ``Tridet: Temporal action detection with relative boundary modeling,'' in \emph{Proceedings of the IEEE/CVF Conference on Computer Vision and Pattern Recognition}, 2023, pp. 18\,857--18\,866.

\bibitem{adatad}
S.~Liu, C.-L. Zhang, C.~Zhao, and B.~Ghanem, ``End-to-end temporal action detection with 1b parameters across 1000 frames,'' \emph{arXiv preprint arXiv:2311.17241}, 2023.

\bibitem{vittad}
M.~Yang, H.~Gao, P.~Guo, and L.~Wang, ``Adapting short-term transformers for action detection in untrimmed videos,'' \emph{arXiv preprint arXiv:2312.01897}, 2023.

\bibitem{adsformer}
Q.~Li, G.~Zu, H.~Xu, J.~Kong, Y.~Zhang, and J.~Wang, ``An adaptive dual selective transformer for temporal action localization,'' \emph{IEEE Transactions on Multimedia}, 2024.

\bibitem{actionmamba}
G.~Chen, Y.~Huang, J.~Xu, B.~Pei, Z.~Chen, Z.~Li, J.~Wang, K.~Li, T.~Lu, and L.~Wang, ``Video mamba suite: State space model as a versatile alternative for video understanding,'' \emph{arXiv preprint arXiv:2403.09626}, 2024.

\bibitem{tsp}
H.~Alwassel, S.~Giancola, and B.~Ghanem, ``Tsp: Temporally-sensitive pretraining of video encoders for localization tasks,'' in \emph{Proceedings of the IEEE/CVF International Conference on Computer Vision}, 2021, pp. 3173--3183.

\bibitem{temporalmaxer}
T.~N. Tang, K.~Kim, and K.~Sohn, ``Temporalmaxer: Maximize temporal context with only max pooling for temporal action localization,'' \emph{arXiv preprint arXiv:2303.09055}, 2023.

\end{thebibliography}

\vfill

\end{document}